\documentclass[times,twocolumn,final]{elsarticle}

\usepackage{medima}
\usepackage{framed,multirow}

\usepackage{verbatim}

\usepackage{amssymb}
\usepackage{latexsym}
\usepackage{hyperref}
\usepackage{color}
\usepackage{amsmath}
\usepackage{booktabs} 
\usepackage{subcaption}
\usepackage{graphicx}
\usepackage{comment}

\usepackage{url}
\usepackage{xcolor}

\usepackage{hyperref}

\definecolor{newcolor}{rgb}{.8,.349,.1}

\journal{Medical Image Analysis}

\begin{document}

\verso{Z. Zhu \textit{et~al.}}

\begin{frontmatter}

\title{Boosting Knowledge Diversity, Accuracy, and Stability via Tri-Enhanced Distillation for Domain Continual Medical Image Segmentation }%

\author[1]{Zhanshi \snm{Zhu}}
\author[1]{Xinghua \snm{Ma}}
\author[2]{Wei \snm{Wang}\corref{cor1}}
\ead{wangwei2019@hit.edu.cn}

\author[5]{Suyu \snm{Dong}}
\author[1]{Kuanquan \snm{Wang}\corref{cor1}}
\ead{wangkq@hit.edu.cn}

\author[6]{Lianming \snm{Wu}}
\author[1]{Gongning \snm{Luo}\corref{cor1}}
\ead{luogongning@hit.edu.cn}

\author[5]{Guohua \snm{Wang}}

\author[4]{Shuo \snm{Li}}

\cortext[cor1]{Corresponding author.}

\address[1]{Faculty of Computing, Harbin Institute of Technology, Harbin, China}
\address[2]{Faculty of Computing, Harbin Institute of Technology, Shenzhen, China}
\address[5]{College of Computer and Control Engineering, Northeast Forestry University, Harbin, China}
\address[6]{Department of Radiology, Renji Hospital, School of Medicine, Shanghai Jiao Tong University, Shanghai, China}
\address[4]{Department of Biomedical Engineering, Case Western Reserve University, Cleveland, OH 44106 USA}

\received{** *** ****}
\finalform{** *** ****}
\accepted{** *** ****}
\availableonline{** *** ****}

\begin{abstract}
Domain continual medical image segmentation plays a crucial role in clinical settings. This approach enables segmentation models to continually learn from a sequential data stream across multiple domains. However, it faces the challenge of catastrophic forgetting. Existing methods based on knowledge distillation show potential to address this challenge via a three-stage process: distillation, transfer, and fusion. Yet, each stage presents its unique issues that, collectively, amplify the problem of catastrophic forgetting. 
To address these issues at each stage, we propose a tri-enhanced distillation framework. (1) \textbf{Stochastic Knowledge Augmentation} reduces redundancy in knowledge, thereby increasing both the diversity and volume of knowledge derived from the old network. (2) \textbf{Adaptive Knowledge Transfer} selectively captures critical information from the old knowledge, facilitating a more accurate knowledge transfer. (3) \textbf{Global Uncertainty-Guided Fusion} introduces a global uncertainty view of the dataset to fuse the old and new knowledge with reduced bias, promoting a more stable knowledge fusion. 
Our experimental results not only validate the feasibility of our approach, but also demonstrate its superior performance compared to state-of-the-art methods. We suggest that our innovative tri-enhanced distillation framework may establish a robust benchmark for domain continual medical image segmentation.

\end{abstract}

\begin{keyword}
\MSC \\
41A05  \\
41A10  \\
65D05  \\
65D17 \\
\KWD \\
Domain Continual Learning \\
Medical Image Segmentation \\
Knowledge Distillation \\

\end{keyword}

\end{frontmatter}


\section{Introduction}
\begin{figure*}
	\centering
	\includegraphics[width=\linewidth]{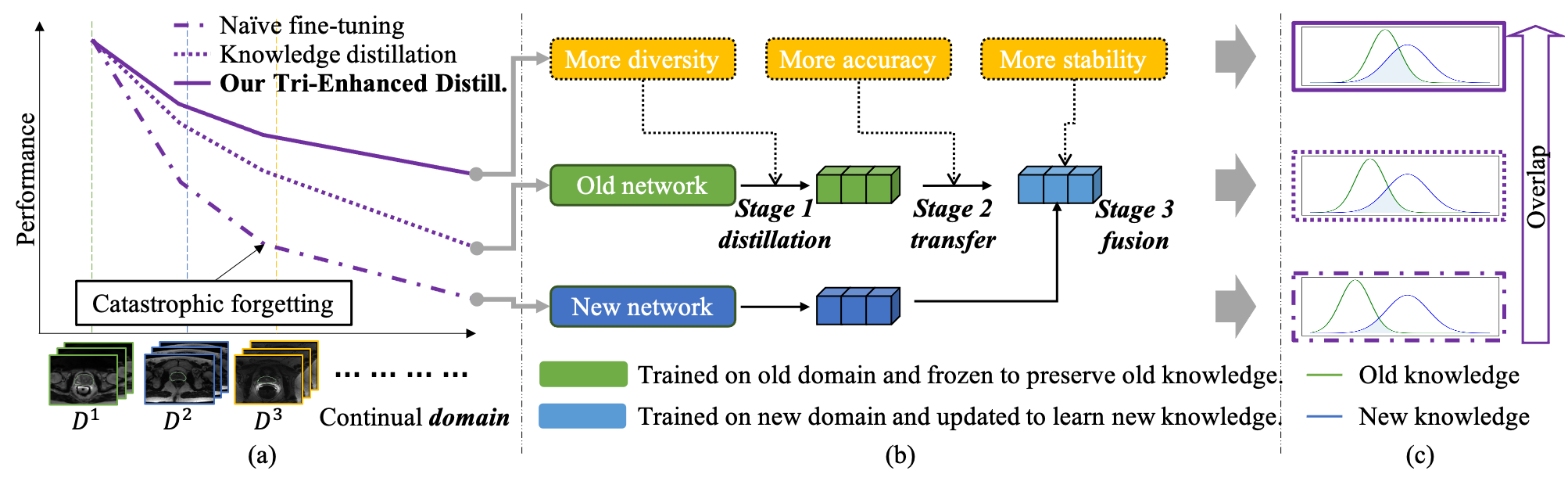}
	\caption{
		(a) Catastrophic forgetting is a significant challenge in domain continual medical image segmentation. (b) Our tri-enhanced distillation framework boosts diversity, accuracy, and stability of knowledge at each stage of the distillation process. (c) Knowledge distillation-based methods help increase the overlap between old and new knowledge, thereby mitigating catastrophic forgetting.}
	\label{cf}
\end{figure*}
\begin{figure}[h]
	\centering
	\includegraphics[width=0.50\textwidth]{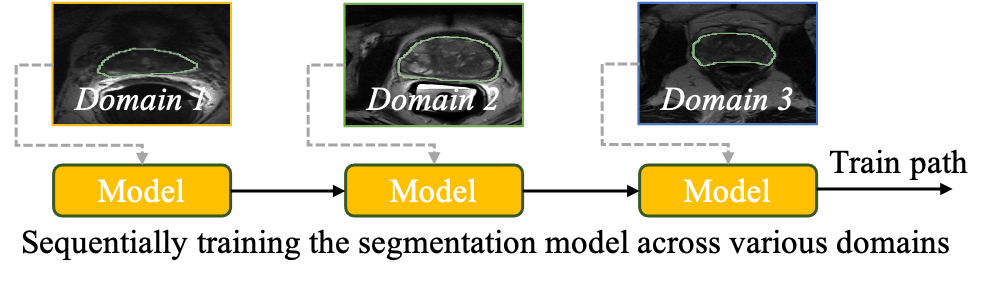}
	\caption{Illustration of domain continual medical image segmentation.} 
	\label{cl}
\end{figure}

Domain continual medical image segmentation holds significant importance in clinical scenarios, despite the challenge of catastrophic forgetting. It involves the sequential reception of data from different domains, thereby enabling a segmentation model to learn without the need to store or access data from previous domains (Fig \ref{cl}). 
Medical image segmentation plays a critical role by accurately identifying and localizing specific structures or areas of interest within medical images, thus providing indispensable information for clinical decision-making \citep{litjens2017survey}. Furthermore, domain continual learning brings a notable impact to the field of medical image segmentation, mainly due to its two key advantages \citep{pianykh2020continuous, van2019three}:
(1) Firstly, domain continual learning enables the quick and efficient integration of new imaging modalities or datasets \citep{gonzalez2020wrong}. With the constant emergence of new data in medical applications, domain continual learning ensures the model continually learns from each new dataset, thereby progressively improving its performance. Unlike naive fine-tuning, which merely adjusts a pre-trained model to a specific new dataset and often necessitates complete retraining to maintain performance on older domains, domain continual learning not only adapts to new datasets but also strategically retains older datasets. Consequently, the domain continual learning approach outperforms naive fine-tuning in terms of training speed and resource efficiency, eliminating the need for extensive model retraining.
(2) Secondly, domain continual learning helps overcome challenges posed by privacy regulations and storage restrictions. These factors limit data availability and complicate the combination of data from different sources \citep{zhang2021comprehensive}. Despite these limitations, domain continual learning ensures the integration of data from various sources through sequential training, allowing the model to adapt to a diverse array of medical situations.
However, domain continual medical image segmentation faces a significant challenge: catastrophic forgetting (Fig. \ref{cf} (a)) \citep{mccloskey1989catastrophic}. This phenomenon is characterized by a model's deteriorating performance on previously learned domains when trained on new ones.

\begingroup
\renewcommand{\thefootnote}{}
\footnotetext{
	\copyright~2024. This manuscript version is made available
	under the CC BY-NC-ND 4.0 license
	(\url{https://creativecommons.org/licenses/by-nc-nd/4.0/}).
	The final published version is available at
	\url{https://doi.org/10.1016/j.media.2024.103112}.
}
\endgroup

To mitigate catastrophic forgetting, knowledge distillation-based methods have emerged as potential solutions \citep{cermelli2020modeling, douillard2021plop, yang2022uncertainty, hinton2015distilling, li2020continual}. 
In these methods, the old network, serving as the 'teacher', maintains its fixed parameters, while the new 'student' network learns from the teacher's accumulated knowledge. During knowledge distillation, the student learns the teacher's outputs, effectively absorbing its knowledge. This ensures the teacher's knowledge distribution stays unchanged, as the student network focuses on learning the teacher's expertise without altering its knowledge base. The procedure of these distillation-based methods can be delineated into three stages, as illustrated in Fig. \ref{cf} (b): distillation of knowledge from the old network using current domain data, transferring this old knowledge into the new network, and finally, fusing the old and new knowledge, often through a guidance by uncertainty, to achieve balance within the new network. As shown in Fig. \ref{cf} (c), knowledge distillation minimizes the difference between old and new knowledge, thereby increasing their overlap and reducing potential conflicts, and aiding in mitigating catastrophic forgetting. Hence, these distillation-based methods show considerable potential in maintaining model efficiency and adaptability for a range of applications, including medical image segmentation.

\begin{figure*}[h]
	\centering
	\includegraphics[width=\textwidth]{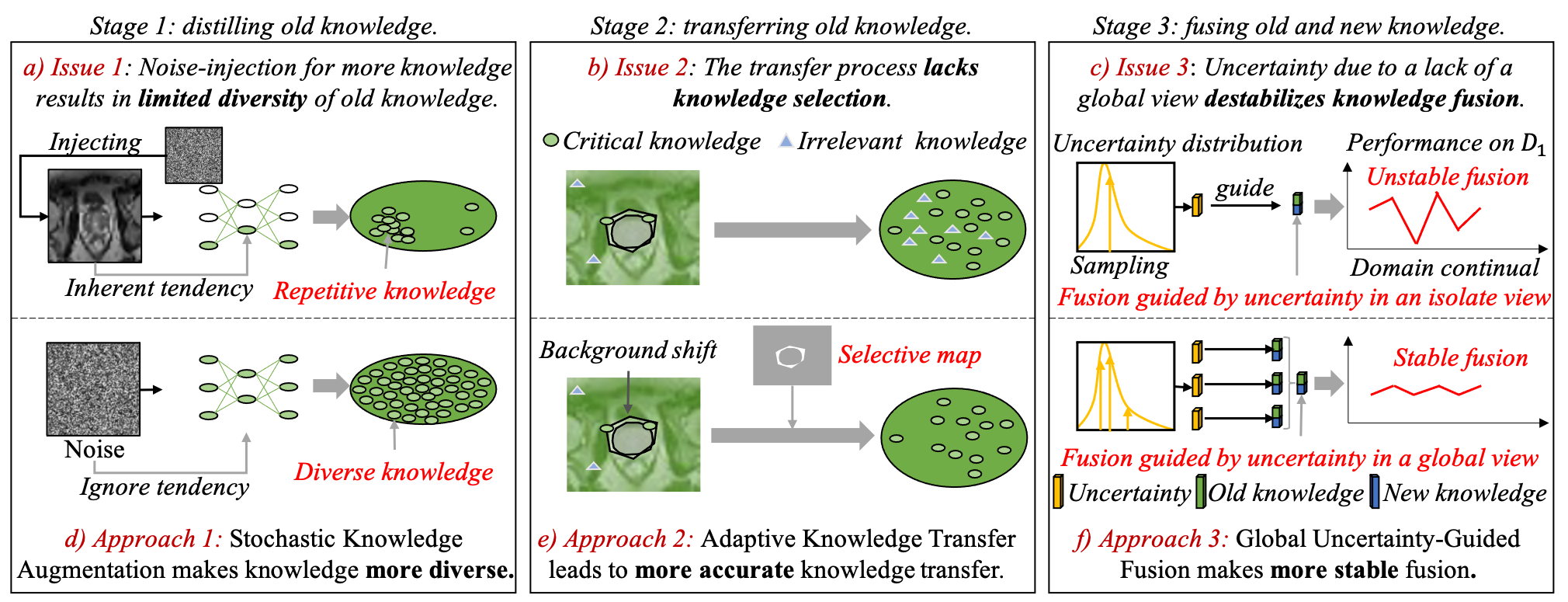}
	\caption{Our tri-enhanced distillation framework refines existing knowledge distillation-based methods. It increases the diversity of distilled knowledge, enhances the accuracy of transferred knowledge, and ensures stability during knowledge fusion.} 
	\label{contributions}
\end{figure*}

However, existing knowledge distillation-based methods face issues at each stage - distillation, transfer, and fusion - which cumulatively hamper the mitigation of catastrophic forgetting.
(1) The first issue (Fig. \ref{contributions} (a)) is the limited diversity and quantity of the distilled old knowledge. Given the limited availability of images in new domains, achieving full distillation of old knowledge remains challenging. Due to the inherent distribution of the current domain, the distilled knowledge tends to be repetitive and less diverse, even when noise injection strategies are employed for data augmentation \citep{chlap2021review}.
(2) The second issue (Fig. \ref{contributions} (b)) is the lack of transfer selection in current distillation-based methods. The segmentation probability map, generated from the old network, is treated as old knowledge. However, the value of its regions to the new network can vary significantly \citep{li2022knowledge,wang2023improved}. Absence of a knowledge selection mechanism can result in the transfer of irrelevant knowledge to the new network. Therefore, some critical knowledge is inadequately focused, such as background domain shift - a major cause of catastrophic forgetting in continual segmentation learning. 
(3) The third issue (Fig. \ref{contributions} (c)) is unstable knowledge fusion, referring to the significant fluctuation in the model's performance during the process of domain continual medical image segmentation. Knowledge fusion is typically achieved by learning old and new segmentation probability maps simultaneously. Uncertainty is utilized to facilitate knowledge fusion, aiming to balance old and new knowledge at the pixel-wise segmentation level \citep{ douillard2021plop}. However, the uncertainty derived from an individual sample may be biased due to an isolated view in datasets, resulting in unstable knowledge fusion during domain continual segmentation. 

We propose a tri-enhanced distillation framework specifically designed to address above issues. As shown in Fig. \ref{cf} (b), it improves the diversity, accuracy, and stability of knowledge at each stage, thereby effectively mitigating catastrophic forgetting.
Firstly (Fig. \ref{contributions} (d)), we propose a stochastic knowledge augmentation approach to distill more diverse old knowledge. 
This approach utilizes noise images generated with pixel values randomly sampled from a Gaussian distribution to extract richer knowledge from the old network. With this method, the inherent distribution characteristics of the current data are ignored, thus expanding the range of knowledge acquisition and achieving a diversified distillation of old knowledge.
Secondly (Fig. \ref{contributions} (e)), we propose an adaptive knowledge transfer approach to improve accuracy of transferred knowledge. This method uses a selective map to filter old knowledge and captures critical domain shift information between the old and new knowledge. By emphasizing this critical knowledge at the pixel-wise segmentation level, our approach enables a more accurate transfer of old knowledge.
Finally (Fig. \ref{contributions} (f)), we propose a global uncertainty-guided fusion approach to enhance stability in the process of continual learning. This method integrates a global view of dataset uncertainty, consolidating segmentation uncertainties from all batch samples into a unified measure. This uncertainty guides knowledge fusion to enhance stability by enabling mutual reinforcement among similar segmentation targets and introducing controlled diversity for varied targets. These strategies together ensure robust and stable fusion of old and new knowledge.

In summary, our tri-enhanced distillation framework effectively addresses the issues faced in the stages of distillation, transfer, and fusion. This study makes significant contributions for domain continual medical image segmentation as follows:

$\bullet$ A tri-enhanced distillation framework is proposed to boost diversity, accuracy, and stability in distinct stages of the distillation process, aiming to mitigate catastrophic forgetting.

$\bullet$ Stochastic knowledge augmentation is proposed to distill a more diverse range of old knowledge. This method uses noise images to uniformly distill knowledge from the old network.

$\bullet$ Adaptive knowledge transfer is proposed to enable a more accurate knowledge transfer, thereby minimizing the influence of irrelevant elements. This strategy selectively captures the critical information among the old knowledge. 

$\bullet$ Global uncertainty-guided fusion is proposed to facilitate a more stable knowledge fusion process. This method introduces a global view into uncertainty, mitigating biases that arise from individual samples.

The code can be accessed at the following address:
\href{https://github.com/PerceptionComputingLab/TED_DCMIS}{$https://github.com/PerceptionComputingLab/TED\_DCMIS$}

\section{Related works}

In this section, we first provide an overview of recent advances in medical image segmentation, particularly focusing on multi-center learning. We also review the latest studies in continual learning and their relevance to our medical scenario. Finally, we introduce the specific advances and challenges in continual learning methods applied to medical image segmentation.

\subsection{Medical image segmentation}

Medical image segmentation has significantly progressed with the adoption of Fully Convolutional Networks (FCNs) and the integration of data from multiple centers. FCNs are widely used for medical image segmentation \citep{long2015fully}, with most methods \citep{li2018h,zhou2018unet++,isensee2021nnu,tang2022self} adopting encoder-decoder architectures, such as U-Net \citep{ronneberger2015u}. 
One substantial challenge in this field is multi-center learning, where data from various institutions are amalgamated to enhance model performance. This challenge is met with diverse strategies like domain adaptation \citep{guan2021domain}, transfer learning \citep{yu2022transfer}, federated learning \citep{nguyen2022federated}, and continual learning \citep{9915459}. Among these, continual learning is distinguished for its advantages in privacy and efficient resource utilization. It allows for sequential training on varied center data without the need for revisiting previous datasets.
Delving deeper, domain continual learning emerges as a primary scenario among the continual learning contexts, as illustrated by \citet{van2019three}. Within medical domains, the prominence of domain continual learning is unmistakable. This is largely attributed to the inherent challenges posed by variations across data domains from multiple centers \citep{gonzalez2020wrong}.
Inspired by these insights, we introduce a novel domain continual learning approach for medical image segmentation in this study.

\subsection{Continual learning}
Existing continual learning methods to mitigate catastrophic forgetting can be broadly categorized into four groups based on how they store and extract old information: rehearsal-based, isolation-based, regularization-based, and distillation-based methods \citep{zhang_s3r_2023}.

Rehearsal-based methods retain a few samples or generate composite images or features to prevent forgetting old tasks \citep{rebuffi2017icarl, shin2017continual}. Most of these methods train the model by mixing old samples or representations with new task data to mitigate catastrophic forgetting. However, these methods face challenges such as potential privacy violations and memory constraints, as they need to store some old samples and free up memory for the removal of less important ones.

Isolation-based methods employ task-specific parameters to incorporate task-related knowledge. One type of method assigns additional parameters to handle future tasks, mitigating catastrophic forgetting \citep{rusu_progressive_2022}. However, this approach may lead to increased parameter counts, which can limit the model's learning ability as the number of tasks grows. Another isolation-based method controls the number of parameters by pre-defining the network size and training pruning \citep{mallya2018packnet,mallya_piggyback_2018}. However, this method also has an upper limit on the model's learning ability.

Regularization-based methods estimate each network parameter's importance and penalize the model parameter accordingly. For example, Elastic Weight Consolidation (EWC \citep{baweja_towards_2018}) uses the diagonal approximation of Fisher's information matrix to calculate importance parameters. Memory Aware Synapses (MAS \citep{aljundi2018memory}) calculates the weight online by accumulating the sensitivity of the learning function, which is the magnitude of the gradient. However, these methods have deficits in their anti-forgetting mechanism, which leads to decreased plasticity in learning new knowledge \citep{ahn2019uncertainty}.

Distillation-based methods distill the old knowledge and transfer it to new models \citep{li2017learning, li2020continual}. These methods have been shown to be effective and advantageous in continual learning of medical images. Compared with rehearsal-based methods, distillation-based methods require no additional storage and avoid privacy invasion. Compared with isolation-based methods, they have a higher upper limit of learning ability because they do not limit the number of active parameters in the network to train the new data. Compared with regularization-based methods, they guide the learning of the new model according to the knowledge of the old model without reducing its learning ability on the new data. 

Given the privacy constraints of medical images and the plasticity of the model itself, knowledge distillation is a good option to mitigate catastrophic forgetting in domain continual segmentation learning. However, knowledge distillation-based methods encounter challenges at each stage of the distillation process, cumulatively impacting their effectiveness in mitigating catastrophic forgetting. To address these challenges, we propose a novel framework for improved mitigation of catastrophic forgetting

\subsection{Segmentation on domain  continual learning}

Continual learning methods have been applied to medical image analysis, but their application to domain continual segmentation remains limited. Although numerous studies have employed continual learning techniques for medical image analysis, the focal point has largely been on medical image classification \citep{yang2021continual, bayasi2021culprit, ayromlou_class_2022, ravishankar_feature_2019, derakhshani_lifelonger_2022}. Existing methods for segmentation often require extra storage \citep{perkonigg_dynamic_2021, li_domain_incremental_2023} or use regular constraints \citep{zhang2021comprehensive, ozgun2020importance, baweja_towards_2018}, which can invade the privacy or limit the model’s ability of learning. 
In this study, an enhanced distillation framework is proposed, which employs knowledge distillation to mitigate catastrophic forgetting in domain continual segmentation learning.

\begin{figure*}[t]
	\centering
	\includegraphics[width=\textwidth]{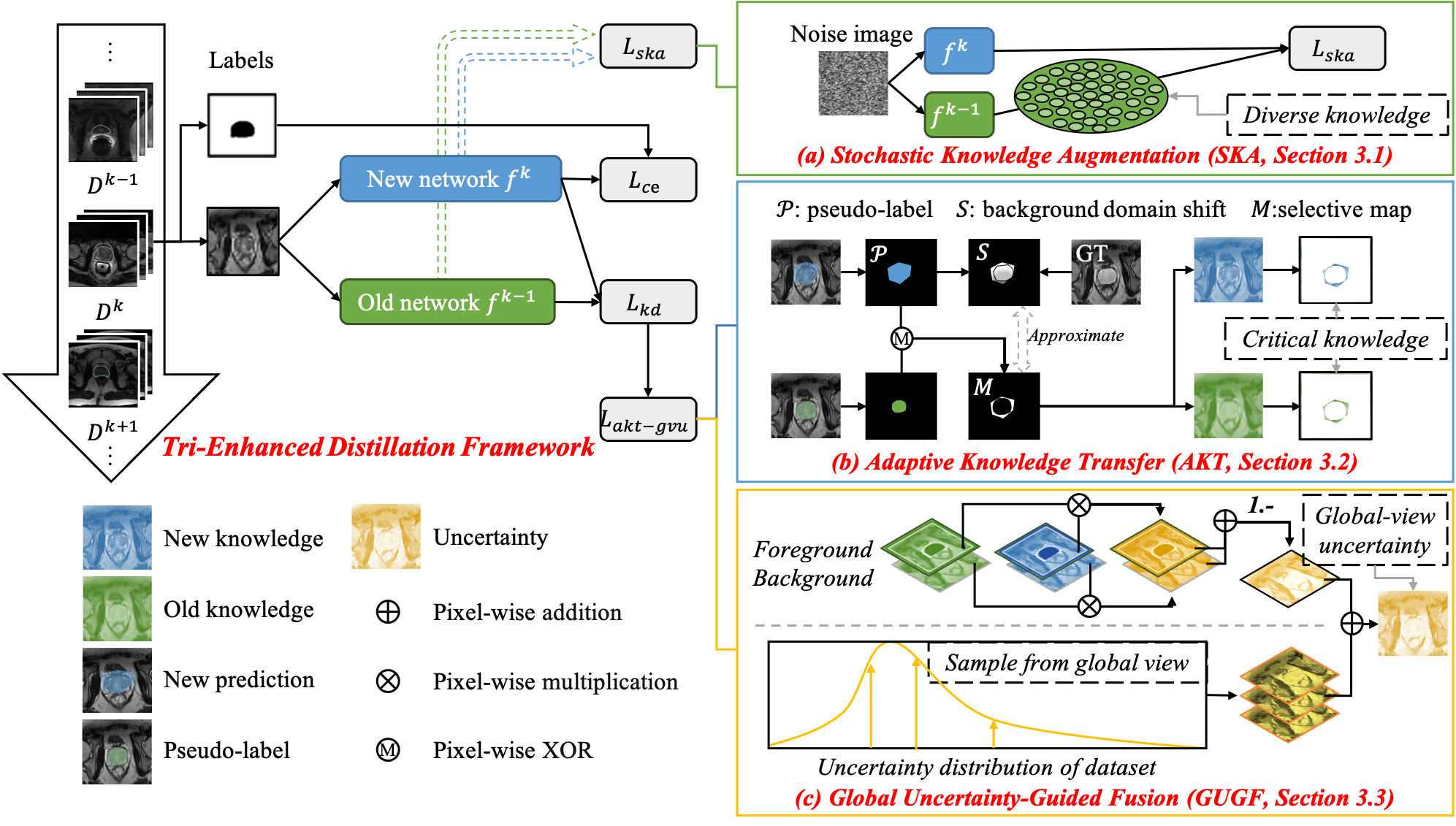}
	\caption{By employing SKA to increase the diversity of distilled knowledge, AKT to enhance the accuracy of transferred knowledge, and GUGF to ensure stability during knowledge fusion, our tri-enhanced distillation framework effectively mitigates catastrophic forgetting.} 
	\label{ed+}
\end{figure*}

\section{Method}

Our tri-enhanced distillation framework comprises three key components that collectively bolster knowledge diversity, accuracy, and stability, as shown in Fig. \ref{ed+}. These components contribute to achieving superior anti-forgetting performance. 
The first component, stochastic knowledge augmentation, employs noise images as additional images to distill a more diverse set of old knowledge. The second component, adaptive knowledge transfer, selectively filters old knowledge using selective maps, thus transferring critical accurate information. The third component, global uncertainty-guided fusion, introduces a global view of datasets into uncertainty to guide stable knowledge fusion between old and new knowledge. 

In domain continual medical image segmentation, the model is sequentially trained on $K$ different domain datasets ($1, \dots, K$) with the objective of segmenting identical semantic targets. Once a new dataset is introduced, the model does not have access to the previous ones. Each of these $K$ datasets includes images and their corresponding targets $(x^k, y^k) \in D^k$ in dataset $k$. In the distillation-based approach, an old trained network serves as a teacher model that offers knowledge into a new student model trained on a new domain dataset $D^k$. The old model is denoted as $f^{k-1}$ and has parameters that remain fixed during training, while the new model, denoted as $f^{k}$, learns new knowledge while consolidating the old ones.

Overall, our tri-enhanced distillation loss is defined in Eq. \ref{ed} and consists of four terms. The first term, $\mathcal{L}_{seg}$ (Eq. \ref{seg}), represents the segmentation loss, which is applied to facilitate the learning of new knowledge. The subsequent three terms contribute to the consolidation of previous knowledge. The term $\mathcal{L}_{kd}$ (Eq. \ref{kdl}) is a standard knowledge distillation loss and used to retain old knowledge. $\mathcal{L}_{ska}$ (Eq. \ref{ska}) is the stochastic knowledge augmentation loss to distill more quantity and diversity of old knowledge. Lastly, $\mathcal{L}_{akt-gvu}$ (Eq. \ref{akt-gvu}) combines the adaptive knowledge transfer with the global uncertainty-guided fusion, which accurately transfer old knowledge and concurrently fuse old and new knowledge in a stable manner during continual learning. The hyper-parameters $\lambda_1$, $\lambda_2$ and $\lambda_3$ are used to maintain a balance among these sub-losses. 

\begin{equation}
	\label{ed}
	\begin{split}
		\mathcal{L}_{ted} = \mathcal{L}_{seg} 
		+ \lambda_1 \mathcal{L}_{kd}  
		+ \lambda_2 \mathcal{L}_{ska}
		+ \lambda_3 \mathcal{L}_{akt-gvu}
	\end{split}
\end{equation}
For all pixels $\mathcal{I}$ of an image $x^k\in D^k$, the segmentation loss $\mathcal{L}_{seg}$ is defined as: 
\begin{equation}\label{seg}
	\mathcal{L}_{seg}(y^k, x^k, f^{k}) = -\frac{1}{|\mathcal{I}|}\sum_{i\in \mathcal {I}} \sum_{c}
	(y_i^k\log(f_c^{k}(x_i^k)))
\end{equation}
where $y_i^k$ denotes the true label for the pixel $x^k_i$. The model $f_c$ generates the probability for class $c \in \mathcal{C}$, where the set $\mathcal{C}$ includes all segmentation targets, such as foreground ($c=1$) and background ($c=0$). 
Additionally, the knowledge distillation loss $\mathcal{L}_{kd}$ in domain continual medical image segmentation for all pixels $\mathcal{I}$ of an image $x^k\in D^k$ is defined as:

\begin{equation}\label{kdl}
	\mathcal{L}_{kd}(f^{k-1}, x^k, f^{k}) 
	= -\frac{1}{|\mathcal{I}|}\sum_{i\in \mathcal {I}} \sum_{c \in \mathcal{C}}
	(f_c^{k-1}(x_i^k)\log(f_c^{k}(x_i^k)))
\end{equation}
This loss function is a form of response-based knowledge distillation, focusing specifically on the pixel-level outputs of the models. Here, $\mathcal{L}_{kd}$ directly compares the predictions of the teacher model $f^{k-1}$ and the student model  $f^{k}$ at each pixel $i$ for every class $c$. This approach is particularly suited for pixel-wise tasks like image segmentation, where capturing fine-grained local differences is crucial. Unlike KL divergence, which is commonly used for assessing the difference in overall probability distributions, our pixel-wise approach allows for a more detailed comparison at the pixel level, thereby facilitating a more targeted and effective knowledge transfer in segmentation scenarios.

\subsection{Stochastic knowledge augmentation for more diversity}
Our SKA (Fig. \ref{ed+} (a)) leverages noise images to uniformly distill knowledge from the old network, facilitating greater diversity and a larger amount of old knowledge. This approach effectively addresses the issue of limited diversity and quantity of old knowledge, which arises from the scarcity of medical images. (1) Firstly, a larger amount of old knowledge is distilled by utilizing noise images. These noise images are randomly generated from a Gaussian distribution, denoted as $\overline{x} \sim \mathcal{N}(0,1)$. This approach allows for the generation of an essentially limitless number of synthetic images. Compared to the limited amount of new domain data, the utilization of noise images enables a larger number of knowledge to be distilled from the old network. (2) Moreover, the diversity of old knowledge is enhanced by conducting Monte Carlo sampling within the space of old knowledge. As the images are randomly sampled from a Gaussian distribution, the additional old knowledge displays a uniform pattern. Utilizing noise as surrogate images eliminates inherent tendency of distribution from the current domain, resulting in a more diverse collection of old knowledge. The loss function for the stochastic knowledge augmentation is defined as follows:

\begin{equation}
	\label{ska}
	\mathcal L_{ska}(f^{k-1} ,\overline{x}^k_i,f^{k})) 
	= -\frac{1}{|\mathcal{I}|}\sum_{i\in \mathcal {I}} \sum_{c \in \mathcal{C}}
	 ( f_c^{k-1} (\overline{x}^k_i) \log(f_c^{k} (\overline{x}^k_i)))
\end{equation}

The model’s performance will be not adversely affected with the inputs of noise images in SKA. As described in Eq. \ref{kdl}, the knowledge distillation (KD) loss with inputs of images is used to ensure the relevance of the distilled knowledge to the image distribution. This KD loss allows SKA to enhance data diversity through noise introduction without risking the acquisition of irrelevant knowledge. Consequently, the SKA is effective in distilling meaningful knowledge even from noise images that significantly differ from the regular distribution.

\textbf{Discussion of the innovation: } Our SKA extends the diversity of the original dataset beyond what is typically achieved with traditional data augmentation (TDA) methods. TDAs tend to produce data within the inherent distribution of the original dataset. In contrast, Gaussian processing generates data that is not confined to the specific features or patterns of the original dataset, thus creating features that lie outside the current dataset's distribution. This approach leads to a broader variety of sample variability, which enhances the dataset's diversity beyond the limitations of traditional augmentation.
Therefore, SKA significantly improves the diversity of the distilled knowledge, effectively addressing the issue of knowledge repetitiveness in distillation-based domain continual medical image segmentation.

\subsection{Adaptive knowledge transfer for more accuracy }

Our AKT (Fig. \ref{ed+} (b)) selectively captures critical information among the old knowledge, enabling a more accurate knowledge transfer. The critical information we focus on is the shift in the background domain, which can be effectively handled by pseudo-label strategies, as shown by PLOP \citep{douillard2021plop} in class incremental segmentation and other domain adaptation studies \citep{chen2021source,9832940,9746286}.
Following the pseudo-label strategies, the background domain shift is quantified by calculating the difference between the pseudo-label and the true label ($y^k \in \mathcal{C}$) of the image $x^k$. The pseudo-labels $\mathcal{P}$ are produced by the old model, with their values determined by $argmax(f^{k-1}(x^k_i)) \in \mathcal{C}$. Function $\mathcal{S}$ is defined to the background domain shift through the equation:

\begin{equation}
	\label{bs}
	\mathcal{S} = y^k_i \oplus  \mathcal{P}_i
\end{equation}
where $\oplus$ signifies an XOR operation and $i$ represents the $i$th pixel in image $x^k$.

For more accurate knowledge transfer, our AKT uses a selective map $\mathcal{M}$ that filters out irrelevant information while retaining critical details represented by $\mathcal{S}$. The selective map is determined by the misclassification between the pseudo-label and output of the new network. On the one hand, $\mathcal{M}$ can be effective in reflecting the key information in $\mathcal{S}$, as the new model more accurately identifies the true labels. This is because that the approximation between $\mathcal{M}$ and $\mathcal{S}$ increases as the difference between the new network's output ($argmax(f^k(x^k))$) and the true label $y^k$ decreases. On the other hand, $\mathcal{M}$ can capture the difference between old and new knowledge, which is also considered critical information for transfer, similar to $\mathcal{S}$. In other words, the selective map treats the conflict between old and new knowledge as the critical information, which is used to improve the accuracy of knowledge transfer. 
Therefore, the value of selective map $\mathcal{M}$ is defined as follows:

\begin{equation}
	\label{mbs}
	\phi(x^k_i, f^k, f^{k-1}) = argmax(f^{k}(x^k_i)) \oplus \mathcal{P}_i
\end{equation}

Based on the selective map (Eq. \ref{mbs}) and the pseudo-label $\mathcal{P}$, the loss for our adaptive knowledge transfer is defined as:

\begin{equation}
	\label{akt}
	\begin{split}
	\mathcal{L}_{akt}(f^{k-1}, x^k, f^{k}) 
	= -\frac{1}{|\mathcal{I}|}\sum_{i\in \mathcal {I}} \phi(x^k_i, f^k, f^{k-1}) \\
	\sum_{c \in \mathcal{C}}
	\mathcal{P}_i\log(f_c^{k}(x_i^k)))
	\end{split}
\end{equation}
where the value $\phi(x^k_i, f^k, f^{k-1}) \in \{0, 1\}$ determines the selective map to capture the background domain shift and the $\sum_{c \in \{0,1\}} \mathcal{P}_i\log(f_c^{k}(x_i^k))$ is a cross entropy to learn the pseudo label of old knowledge.

\textbf{Discussion of the innovation: } Our AKT is a novel selection-based knowledge transfer method. It selectively captures the background domain shift between old and new knowledge, treating this as critical and accurate knowledge. Therefore, it significantly improves the accuracy of transferred knowledge, effectively addressing the issue of transferring irrelevant knowledge in distillation-based domain continual medical image segmentation.

\subsection{Global uncertainty-guided fusion for more stability}

Our GUGF (Fig. \ref{ed+} (c)) mitigates bias that might arise from individual samples, guiding a more stable fusion of old and new knowledge. The stable knowledge fusion refers to a minimal fluctuation in the model's performance during the continual learning process, as illustrated by the line chart in Fig. \ref{contributions} (f). The GUGF incorporates a global view of dataset uncertainty into the learning process. Specifically, GUGF aggregates the segmentation uncertainty of all samples in a current batch into a unified uncertainty measure. This approach bolsters knowledge fusion stability from two angles:
(1) Mutual Reinforcement: When shapes of segmentation targets in a batch are similar, their uncertainties can mutually reinforce each other, fostering a deeper semantic understanding of the current target.
(2) Diversity Introduction: In cases where segmentation targets in a batch differ significantly, the resulting diverse uncertainties can guide the fusion process, akin to introducing noise. This diversity aids in enhancing the robustness and stability of knowledge fusion.
By employing this dual-faceted strategy, the GUGF ensures that the fusion of new and old knowledge is stable.

Typically, uncertainty is used to balance old and new knowledge, mitigating their conflicts to enhance stability \citep{douillard2021plop, yang2022uncertainty}. For instance, consider a scenario where conflicts arise: the edge of an object is labeled as foreground by the pseudo-label, while labeled as background by the ground truth. This conflict can be mitigated under uncertainty guidance. As shown in Fig. \ref{ed+} (c), the GUGF proposes the global-view uncertainty (GVU) to firstly combine old and new predictions as Eq. \ref{un}. This function takes into account the confidence of the new ($p_c=f_c^k(x^k_i)$) and old network ($q_c=f_c^{k-1}(x^k_i)$) in determining the degree of semantic conflict. The product of $p_c$ and $q_c$ reveals the confidence of foreground ($c=1$) and background ($c=0$). Then, $1- \sum_{c\in {0,1}} p_cq_c$ signifies the ambiguity regarding the foreground or background, which can be accounted for uncertainty.
\begin{equation}
	\label{un}
	h(x^k_i,f^k, f^{k-1}) = 1-\sum_{c\in \{0,1\}} f_c^k(x^k_i)f_c^{k-1}(x^k_i)
\end{equation}

To mitigate potential bias in uncertainty derived from individual samples, the global-view uncertainty adapts a global perspective on the dataset's uncertainty. For mutual reinforcement and diversity introduction , the global view is incorporated into the uncertainty by summing a batch of uncertainty maps, defined as follows:
\begin{equation}
	uncertainty_{gvu} =	\sum_{x_i^k \in {batch}} h(x^k_i,f^k,f^{k-1})
\end{equation}

There are several points to consider regarding our global-view uncertainty. (1) Firstly, it is important to note that this uncertainty measures the degree of semantic conflict. Hence, higher uncertainty implies more certainty of a conflict, which necessitates assigning a higher weight to the loss to fight the true labels. (2) Moreover, it is essential to incorporate a global view into individual uncertainty (Eq. \ref{un}). Otherwise, the role of uncertainty in enhancing the stability of knowledge fusion will be limited, since the selective map Eq. \ref{mbs} is almost equivalent to the individual uncertainty Eq. \ref{un} for loss function. Typically, the higher uncertainty indicates potential misclassified regions (the selective map). In this regard, it would be reasonable to utilize either a misclassified-guided binary mask or an uncertainty-guided weight map, rather than both simultaneously. On the contrary, our global-view uncertainty method avoids this pitfall by introducing a global view term. (3) Finally, the background shift-aware transfer and the global-view uncertainty must be employed together. While the former transfers more accurate old knowledge, it also intensifies conflicts between old and new knowledge. Therefore, the global-view uncertainty is necessary to mitigate this conflict. The loss function in Eq. \ref{akt} is modified by integrating global-view uncertainty as follows:

\begin{equation}
	\label{akt-gvu}
	\begin{split}
		\mathcal{L}_{akt-gvu} = - \frac{1}{|\mathcal{I}|} \sum_{i\in\mathcal{I}}  
		\phi(x^k_i, f^k, f^{k-1})  \\
		\sum_{x_i^k \in {batch}} h(x^k_i,f^k,f^{k-1})\\
		\sum_{c \in \mathcal{C}}
		\mathcal{P}_i\log(f_c^{k}(x_i^k))
	\end{split}
\end{equation}

\textbf{Discussion of the innovation: } Our GUGF proposes a novel uncertainty estimation method with a global view of datasets. It samples a batch of uncertainty from the datasets and measures the degree of semantic conflict between old and new knowledge in a global view. Therefore, it can be used to significantly improve the stability of knowledge fusion, effectively addressing the issue of significant fluctuation in the model's performance during the process of distillation-based domain continual medical image segmentation.

\section{Experimental setting}

\subsection{Datasets and preparing}

\begin{table}[h]
	\centering
	\caption{Summary of the segmentation tasks and their corresponding domains. These domains exhibit diversity due to differing acquisition protocols and equipment.}
	\label{dataset}
	\begin{tabular}{cccc}
		\toprule
		Task & Sources & Domain ID & No. \\ 
		\midrule
		\multirow{6}{*}{Prostate}  & \multirow{2}{*}{\citet{bloch2015nci}}  &  RUNMC& 30 \\
		{} &  &  BMC&  30\\
		\cmidrule(lr){2-4}
		{} & \citet{lemaitre2015computer} &  I2CVB&  19\\
		\cmidrule(lr){2-4}
		{} &&  UCL&  13\\
		{} &  \citet{litjens2014evaluation}  &  BIDMC&  12\\
		{} &  &  HK&  12\\
		\midrule
		\multirow{4}{*}{Cardiac} & \multirow{4}{*}{\citet{9458279} }& Siemens & 95  \\
		{}& & Philips&125\\
         {}& & GE&50\\
		{}& & Canon&50\\
		\bottomrule
	\end{tabular}
\end{table}

We evaluate the performance of proposed tri-enhanced distillation framework on two datasets: MRI prostate segmentation and MRI cardiac segmentation. These tasks exhibit diverse domain sources due to differing acquisition protocols and equipment, details of which are outlined in Table \ref{dataset}.

The \textbf{Prostate Dataset} consists of T2-weighted MRI images \citep{liu2020shape}. The training sequence (RUNMC $\rightarrow$ BMC $\rightarrow$ I2CVB $\rightarrow$ UCL $\rightarrow$ BIDMC $\rightarrow$ HK) is determined randomly. All images are resized in the axial plane to $192\times 192$ pixels and normalized to fit within the range $[0, 1]$.

The \textbf{Cardiac Dataset} is derived from the M$\&$Ms Dataset \citep{9458279}, and the segmentation tasks focus on the left ventricular endocardium (LV-endo), epicardium (LV-epi), and the right ventricular endocardium (RV). The training of the model follows a sequential order, i.e., Siemens $\rightarrow$ Philips $\rightarrow$ GE $\rightarrow$ Canon. All images are resized in the axial plane to dimensions of $256\times 256$ pixels and normalized to the range $[0, 1]$.

\subsection{Comparison methods}
We compare proposed tri-enhanced distillation framework (\textbf{TED}) with several state-of-the-art continual learning methods. As a baseline method, we initially adopt a simple sequential fine-tuning (\textbf{SEQ}) scheme with cross-entropy loss $\mathcal L_{seg}$ \ref{seg}. We then consider two classic regularization-based methods: \textbf{EWC} \citep{kirkpatrick2017overcoming} and \textbf{MAS} \citep{aljundi2018memory}, which add a penalty term to the loss function based on the importance of each parameter to previous tasks. Furthermore, we select a basic knowledge distillation method (\textbf{KD} \citep{li2017learning}) with $\mathcal L_{kd}$ \ref{kdl} and two advanced distillation-based continual segmentation methods ( \textbf{MiB} \citep{cermelli2020modeling} and \textbf{PLOP} \citep{douillard2021plop}). These methods are modified to suit our settings \citep{gonzalez2022lifelong}. A model trained on all domain datasets collectively serves as the upper performance bound for all the approaches (\textbf{Upper bound}). 
\begin{table*}[t]
	\centering
	\caption{Comparison study on prostate datasets demonstrates the advantages and effectiveness of our tri-enhanced distillation framework. We report the performance of past (RUNMC, BMC, I2CVB, UCL, BIDMC) and current domains (HK) separately in the last time step. AVG is the average performance on all the domains. BWT shows the degree of forgetting.}
	\label{comparison_prostate}
	\small
	\begin{tabular}{c|cccccc|cc}
		\toprule
		Methods& {RUNMC}  & {BMC} &  {I2CVB}&  {UCL}&  {BIDMC}& {HK(current)}& {AVG}&{BWT}\\
		\midrule
		{}& \multicolumn{8}{c}{Dice Similarity Coefficient$\uparrow$ (rank)} \\
		\cmidrule(r){2-9}
		Upper bound  &  0.887& 0.850 &0.797& 0.870&0.770&0.845&0.837& {/}\\
		\midrule
		SEQ  &  0.697(3)&0.718(2)  &0.451(7) & 0.705(4)&\textbf{0.751(1)}&0.756(4)&0.680(3)&-0.154(4)\\
		
		EWC  &  0.627(6)&0.618(7)  & 0.495(6)& 0.699(6)&0.620(5)&0.705(7)&0.628(7)&-0.186(7)\\
		
		MAS  &  0.655(4)& 0.675(4) & 0.558(5)&0.716(3) &0.617(6)&0.715(6)&0.656(5)&-0.151(3)\\
		
		KD  &  0.728(2)&  0.707(3)&0.647(2) &0.749(2) &0.653(3)&0.768(3)&0.709(2)&-0.105(2)\\
		
		MiB  &  0.634(5)&0.671(5)&\textbf{0.667(1)}&0.678(7)&0.649(4)&0.777(2)&0.680(3)&-0.154(4)\\
		
		PLOP  & 0.607(7)& 0.664(6)&0.576(4)&0.701(5)&0.544(7)&0.721(5)&0.645(6)&-0.166(6)\\
		
		\textbf{TED (ours)} &\textbf{0.845(1)}&\textbf{0.742(1)}&0.632(3)&\textbf{0.823(1)}&0.707(2)&\textbf{0.803(1)}&\textbf{0.759(1)}&\textbf{-0.059(1)}\\
		\midrule
		\midrule
		
		{}& \multicolumn{8}{c}{Hausdorff Distance 95 $\downarrow$ (rank)}\\
		\cmidrule(r){2-9}
		Upper bound  &  12.65& 13.91  &13.75& 15.69&23.56&27.30&17.81&{/}\\
		\midrule
		SEQ  &  32.86(3)&34.51(4)  &45.03(6) & 47.31(6)&26.38(2)&\textbf{14.47(1)}& 33.43(4)& 20.29(3)\\
		
		EWC  &  35.85(4)&38.86(7)  & 53.74(7)& 37.95(3)&31.02(4)&29.76(6)&37.86(6)&23.02(6) \\
		
		MAS  &  36.14(5)& 24.77(2) & \textbf{25.50(1)}&38.57(4) &34.31(5)&21.30(4)&30.10(3)&30.10(7)\\
		
		KD  &  27.95(2)&  29.44(3)&26.54(2) &39.72(5) &34.97(7)&16.49(2)&29.18(2)&15.02(2)\\
		
		MiB  &  46.53(7)&37.53(6)&30.70(4)&48.28(7)&34.63(16)&29.83(8)&37.92(7)&21.31(4)\\
		
		PLOP  & 42.57(6)& 35.35(5)&37.25(5)&37.44(2)&30.20(3)&26.17(5)&34.83(5)&21.31(4)\\
		
		\textbf{TED (ours)} &\textbf{11.94(1)}&\textbf{23.67(1)}&29.35(3)&\textbf{21.16(1)}&\textbf{19.37(1)}&23.14(3)&\textbf{21.44(1)}&\textbf{6.72(1)}\\
		
		\bottomrule
	\end{tabular}
\end{table*}

\begin{table*}[!h]
	\centering
	\small
	\caption{Comparative analysis on cardiac datasets demonstrates the advantages and effectiveness of our tri-enhanced distillation framework. We report the performance of three anatomical structures across all domain datasets at the last time step. The column 'whole' shows the average performance of three anatomical structures. AVG represents average performance, while BWT indicates the degree of forgetting.}
	\label{comparison_cardiac}
	\begin{tabular}{c|cccc|cccc}
		\toprule
		Methods&LV-endo&LV-epi&RV&whole&LV-endo&LV-epi&RV&whole\\
		\midrule
		{}&\multicolumn{4}{c|}{AVG}&\multicolumn{4}{c}{BWT}\\
		\midrule
		& \multicolumn{8}{c}{Dice Similarity Coefficient$\uparrow$ (rank)} \\
		\midrule
		Upper bound  &0.884&0.810&0.853&0.849&{/}&{/}&{/}&{/}\\
		\midrule
		SEQ  &0.844(6)&0.661(7)&0.780(5)&0.761(7)&-0.050(6)&-0.182(7)&-0.093(6)&-0.108(7)\\
		
		EWC &0.846(5)&0.705(6)&0.752(7)&0.768(6)&-0.043(5)&-0.117(6)&-0.108(7)&-0.089(5)\\
		
		MAS  &0.832(7)&0.718(5)&0.803(3)&0.784(5)&-0.059(7)&-0.100(5)&-0.042(2)&-0.089(6)\\
		
		KD  &0.853(4)&0.729(4)&0.805(2)&0.795(3)&-0.035(4)&-0.089(4)&-0.048(3)&-0.057(3)\\
		
		MiB  &0.857(3)&0.733(3)&0.765(6)&0.785(4)&-0.032(3)&-0.081(3)&-0.092(5)&-0.068(4)\\
		
		PLOP &0.864(2)&\textbf{0.748(1)}&0.789(4)&0.800(2)&-0.016(2)&\textbf{-0.065(1)}&-0.065(4)&-0.048(2)\\
		
		\textbf{TED (ours)} &\textbf{0.887(1)}&0.735(2)&\textbf{0.831(1)}&\textbf{0.817(1)}&\textbf{0.015(1)}&-0.075(2)&\textbf{-0.012(1)}&\textbf{-0.024(1)}\\
		\midrule
		\midrule
		{}&\multicolumn{8}{c}{Hausdorff Distance 95$\downarrow$ (rank)}\\
		\midrule
		Upper bound  &8.73&7.79& 19.07&11.86&{/}&{/}&{/}&{/}\\
		\midrule
		SEQ  &20.92(3)&32.04(6)&19.25(4)&24.07(6)&12.59(6)&28.53(7)&4.06(3)&15.06(7)\\
		
		EWC &22.22(6)&32.04(6)&25.92(7)&26.72(7)&8.09(4)&22.03(6)&10.77(7)&13.63(6)\\		
		
		MAS  &28.92(7)&24.31(5)&16.65(2)&23.29(5)&17.17(7)&20.21(5)&0.46(2)& 12.61(5)\\
		
		KD  & 21.26(4)&19.14(2)&19.20(3)&19.86(3)&12.30(5)&11.54(3)&4.06(3)&7.88(3)\\
		
		MiB  &18.88(2)& 21.12(4)&23.29(6)&21.10(4)& 7.19(2)&14.54(4)&7.28(6)&9.67(4)\\
		
		PLOP &21.51(5)&\textbf{15.68(1)}&20.04(5)&19.07(2)&8.06(3)& \textbf{7.48(1)}&6.58(5)&7.37(2)\\
		
		\textbf{TED (ours)} &\textbf{13.91(1)}&20.60(3)&\textbf{16.27(1)}&\textbf{16.93(1)}&\textbf{2.90(1)}& 9.91(2)&\textbf{-2.84(1)}&\textbf{3.32(1)}\\
		\bottomrule	
		
	\end{tabular}
\end{table*}

\subsection{Implementation details}

Our segmentation network utilizes a mirrored encoder-decoder 2D-UNet architecture. All loss weights $\lambda_1$, $\lambda_2$, and $\lambda_3$ are set to $0.001$, which determine the importance of the respective loss components in the final loss computation. 
Initially, we set the learning rate at 0.0002, subsequently adjusting it to 0.0001 with a 0.99 decay rate at each epoch for later datasets. Training on a GeForce RTX 4070 GPU, the network undergoes 50 epochs with a batch size of 4 for each dataset phase. At each phase, the data is divided into training, validation, and testing sets with a ratio of $60:15:25$, and the best-performing model on the validation set is selected for testing. 
For the comparison methods, we used the same configuration parameters as our method, including the backbone of network, learning rate, loss weight, dataset grouping, etc.

\subsection{Metrics}

To assess our model performance, we primarily use the Average Accuracy (AVG) and Backward Transfer (BWT) metrics derived from the Dice Similarity Coefficient (DSC) and the 95$\%$ Hausdorff Distance (HD95). After training our model on dataset $D_i$, we define $DSC_{i,j}$ as the DSC calculated for the dataset $D_j$. The AVG and BWT metrics, particularly significant for continual learning evaluation, are subsequently calculated based on these DSC values. 

AVG represents the mean DSC across all test datasets and is given by

\begin{equation}
	\label{AVG}
	AVG=\frac{1}{K}{\sum_{i=1}^K}DSC_{K,i}
\end{equation}

BWT, on the other hand, quantifies the extent of forgetting. It's expressed as
\begin{equation}
	\label{BWT}
	BWT=\frac{1}{K-1}{\sum_{i=1}^{K-1}}(DSC_{K,i}-DSC_{i,i})
\end{equation}
Negative BWT values indicate a higher degree of forgetting. 

Higher AVG and BWT values suggest improved model performance when the metrics are based on DSC. Conversely, when BWT and AVG are computed from HD95, lower values signal superior model performance.

\section{Result and discussion}

\subsection{Effectiveness in Mitigating Catastrophic Forgetting}

The results presented in Tables \ref{comparison_prostate} and \ref{comparison_cardiac} highlight the capability of our tri-enhanced distillation (TED) framework to mitigate catastrophic forgetting.

\textbf{Prostate Dataset Analysis (Table \ref{comparison_prostate})}: Following the training on the last dataset (HK), it's evident that the naive sequential fine-tuning (SEQ) is significantly impacted by catastrophic forgetting. The BWT displays an average decline of 15.4\% in DICE and an increment of 20.29 in HD95. Conversely, our TED framework considerably reduces this effect. The average DICE performance jumps from 68.0\% to 75.9\% (an improvement of 7.9\%), while the average HD95 drops from 33.43 to 21.44. This leads to a reduction in DICE degradation from 15.4\% to 5.9\% and an improvement in HD95 from 20.29 to 6.72 across all domains. Notably, even though our TED underperforms slightly on the BIDMC dataset in terms of DICE compared to the SEQ baseline, its average DICE performance across all domains shows a marked enhancement. While TED's HD95 performance on the latest trained dataset (HK) lags behind SEQ, this is justifiable given that the HK dataset was extensively trained within SEQ. 

\textbf{Cardiac Dataset Analysis (Table \ref{comparison_cardiac})}: Our TED framework consistently outperforms the sequential training (SEQ) in segmenting the left ventricular endocardium (LV-endo), left ventricular epicardium (LV-epi), and right ventricle (RV). Overall, the average DICE score rises from 76.1\% to 81.7\%, and the average HD95 reduces from 24.07 to 16.93. In the LV-endo segmentation, while SEQ exhibits a 5.0\% decline in DICE and a 12.59 increase in HD95, our TED framework remarkably reduces this forgetting, only showing a 1.5\% decrease in DICE and an increase of 2.90 in HD95. Notably, the BWT for DICE even reflects a +1.5\%, suggesting no loss of knowledge on average. For the LV-epi segmentation, TED demonstrates substantial enhancements over SEQ. The BWT for DICE improves from -18.2\% to -7.5\%, and the BWT for HD95 improves from 28.53 to 9.91.  In the RV segmentation, TED's performance is especially commendable. The DICE decrement narrows from 9.3\% to 1.2\%. Furthermore, the BWT for HD95, at -2.84, implies no knowledge loss on average.

\subsection{Comparison with State-of-the-art Methods}
\textbf{Quantitative Prostate Segmentation:} 
Table \ref{comparison_prostate} underscores the distinct advantage of our tri-enhanced distillation (TED) framework in mitigating catastrophic forgetting. TED consistently achieves the highest average DICE score, indicating superior overlap between the predicted and actual regions, and the lowest HD95, highlighting the precision of segmentation boundaries. This performance underscores TED's ability to harmonize the retention of prior knowledge with the assimilation of new data.

$\bullet$ (1) Performance Analysis of Regularization-based Methods: 
Despite its reputation for countering catastrophic forgetting, EWC's performance lags across domains. Specifically, it trails TED by approximately 13.1\% in average DICE, emphasizing TED's superiority.
Offering only a slight improvement over EWC, MAS still doesn't approach TED's benchmarks, with a DICE score deficit of about 10.3\%. The limited effectiveness of these regularization-based methods likely stems from their inadequate anti-forgetting mechanisms, compromising learning adaptability across new domains \citep{ahn2019uncertainty}.

$\bullet$(2) Delving into Knowledge Distillation and its Variants:
Naive Knowledge Distillation (KD) comes closest to TED in performance. TED edges out KD by approximately 5\% in average DICE and boasts a superior HD95.
While KD does mitigate catastrophic forgetting effectively, its variants PLOP and MiB only offer minor enhancements.  MiB, in particular, excels in the I2CVB dataset, even surpassing TED in DICE score. Yet, when averaged across all domains, it shows similar performance with SEQ, both lagging behind TED by 7.9\%. PLOP, on the other hand, consistently underperforms relative to TED. Despite its competitive DICE scores in certain domains, on average, it trails TED by 11.4\%. Evaluating HD95 metrics, MiB shows varied performance. While it shines on the I2CVB dataset, its average HD95 across domains is notably higher than TED's, suggesting compromised boundary precision. PLOP faces a similar predicament, with its HD95 metrics consistently trailing TED's across most domains, and an average HD95 deficit of 13.39 units compared with TED. Examining BWT for DICE reveals MiB's vulnerability to catastrophic forgetting, mirroring SEQ with a decline of 15.4\%. This is further accentuated by its HD95 BWT, which is inferior to TED's by 14.59 units. This analysis indicates that techniques dependent on distillation for class continual segmentation learning may not be entirely suitable for domain continual segmentation learning. 

$\bullet$ (3) Notably, our proposed TED framework outperforms all other state-of-the-art methods, thus validating the advantages of our approach in domain continual medical image segmentation. In the view of average DICE and HD95, TED achieves 75.9\% and 21.44 units respectively, showing a great improvement compared with other methods. On the other hand, For BWT to measure the forgetting, TED
shows the significant ability for anti-forgetting, achieving the best 5.9\% DICE and 6.72 units HD95 forgetting.

\textbf{Quantitative Cardiac Segmentation:}
The results showcased in Table \ref{comparison_cardiac} are indicative of the prowess of our tri-enhanced distillation (TED) framework across various cardiac anatomical structures. TED consistently delivers top performance, achieving the highest average DICE score and the lowest HD95.

$\bullet$(1) LV-endo Analysis:
For the LV-endo structure, EWC appears to falter both in DICE, trailing TED by a significant 4.1\%, and in HD95, where it's surpassed by TED by 8.31 units. MAS, though slightly better than EWC in DICE with a performance deficit of 5.5\% relative to TED, lags considerably in HD95 by 14.99 units. KD, on the other hand, is closer to TED in DICE, with a difference of about 3.4\%, but has an HD95 that's greater by 7.35 units. In DICE, PLOP outperforms MiB by 0.7\% but both are surpassed by TED, which leads by 3\% and 2.3\% respectively. In terms of HD95, MiB and PLOP trail TED by 4.97 and 7.6 units respectively.

$\bullet$(2) LV-epi Analysis:
In the LV-epi segment, both EWC and MAS fall short of TED by 3\% and 1.7\% in DICE respectively. Their HD95 scores also trail TED by 11.44 and 3.71 units respectively. KD stands out with its performance closely matching TED's in DICE, with a difference of only 0.6\%, but its HD95 score is greater by 1.46 units. PLOP shines in DICE, achieving the top score and even surpassing TED by 1.3\%. However, when it comes to HD95, PLOP trails TED by 0.1 units. MiB's performance, while competitive in DICE, trailing TED by only 0.2\%, has an HD95 score that's 0.52 units greater than TED's.

$\bullet$(3) RV Analysis:
In the RV structure, EWC and MAS exhibit significant shortfalls, with TED outperforming them by 7.9\% and 2.8\% in DICE, and by 9.65 and 0.03 units in HD95, respectively. KD is closely aligned with TED in DICE, showing only a 2.6\% difference, but has an HD95 that's greater by 2.93 units. While MiB and PLOP have competitive DICE scores, trailing TED by 6.6\% and 4.2\% respectively, their HD95 metrics lag behind TED's by 7.02 and 3.77 units respectively.

$\bullet$(4) Overall Analysis:
Considering a holistic performance across all structures (denoted as 'whole'), TED consistently shows superior overall performance compared to other methods examined in this study. EWC and MAS, trail TED by 4.9\% and 3.3\% in average DICE, respectively, and by 9.83 and 10.36 units in HD95. KD, while being a strong contender in DICE, being outdone by only 2.2\%, lags in HD95 by 2.87 units. MiB and PLOP  are overshadowed by TED in both DICE, lagging by 3.2\% and 1.7\% respectively, and in HD95, with deficits of 4.17 and 2.14 units.

$\bullet$(5) In summary, the data in Table \ref{comparison_cardiac} stands as a testament to TED's superiority in domain continual cardiac image segmentation. Its consistent top performance across different cardiac structures accentuates advantage, not only in terms of segmentation accuracy but also in capturing precise boundaries.

\begin{figure*}[!h]
	\centering
	\includegraphics[width=\textwidth]{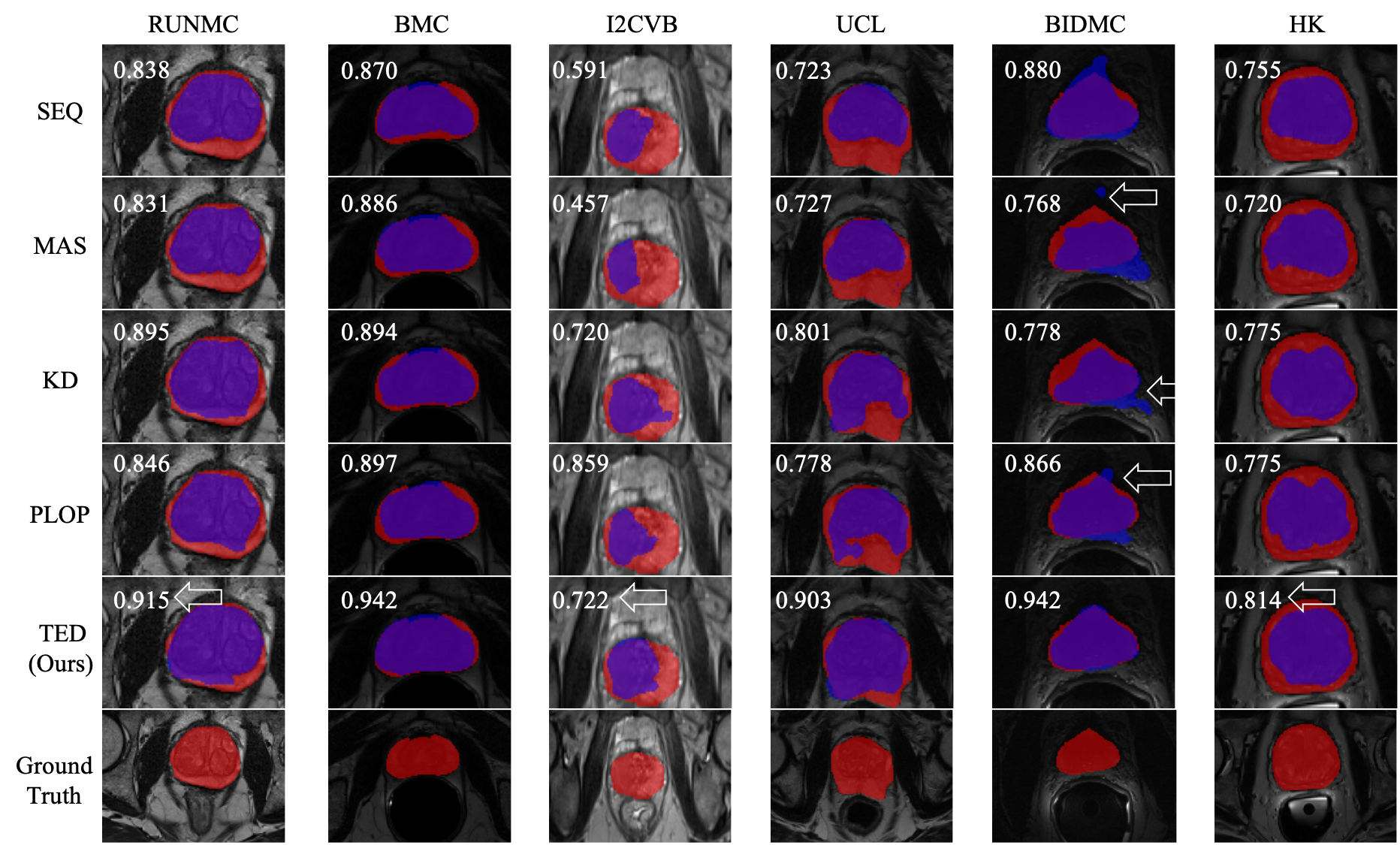}
	\caption{The visualization of results across all domains, post-training on the final domain HK, distinctly underscores the superior of our tri-enhanced distillation framework in prostate segmentation tasks when compared to other methodologies.} 
	\label{result_prostate}
\end{figure*}
\begin{figure*}[!h]
	\centering
	\includegraphics[width=\textwidth]{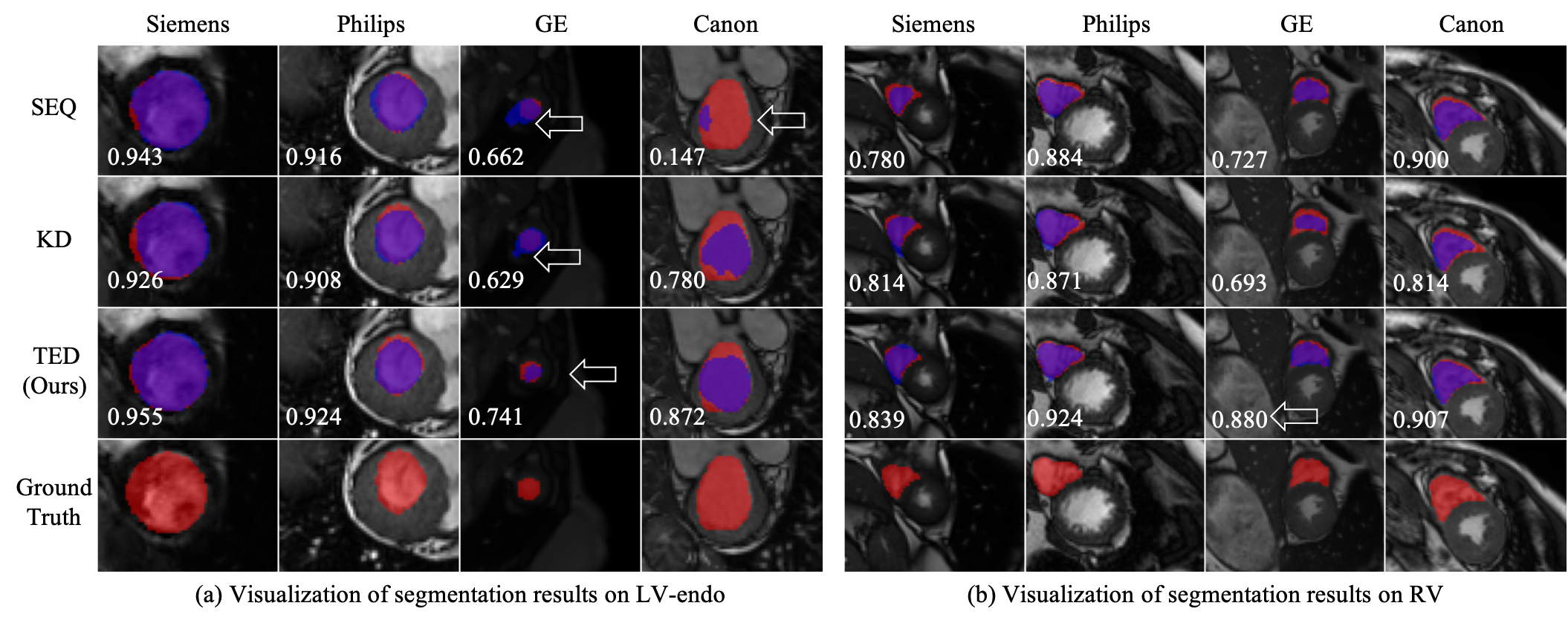}
	\caption{The visualization of segmentation results, after training on the final domain Canon, displays the advanced performance of our tri-enhanced distillation framework in cardiac segmentation tasks relative to other methods.} 
	\label{result_cardiac}
\end{figure*}
\begin{figure*}[!h ]
	\centering
	\includegraphics[width=\textwidth]{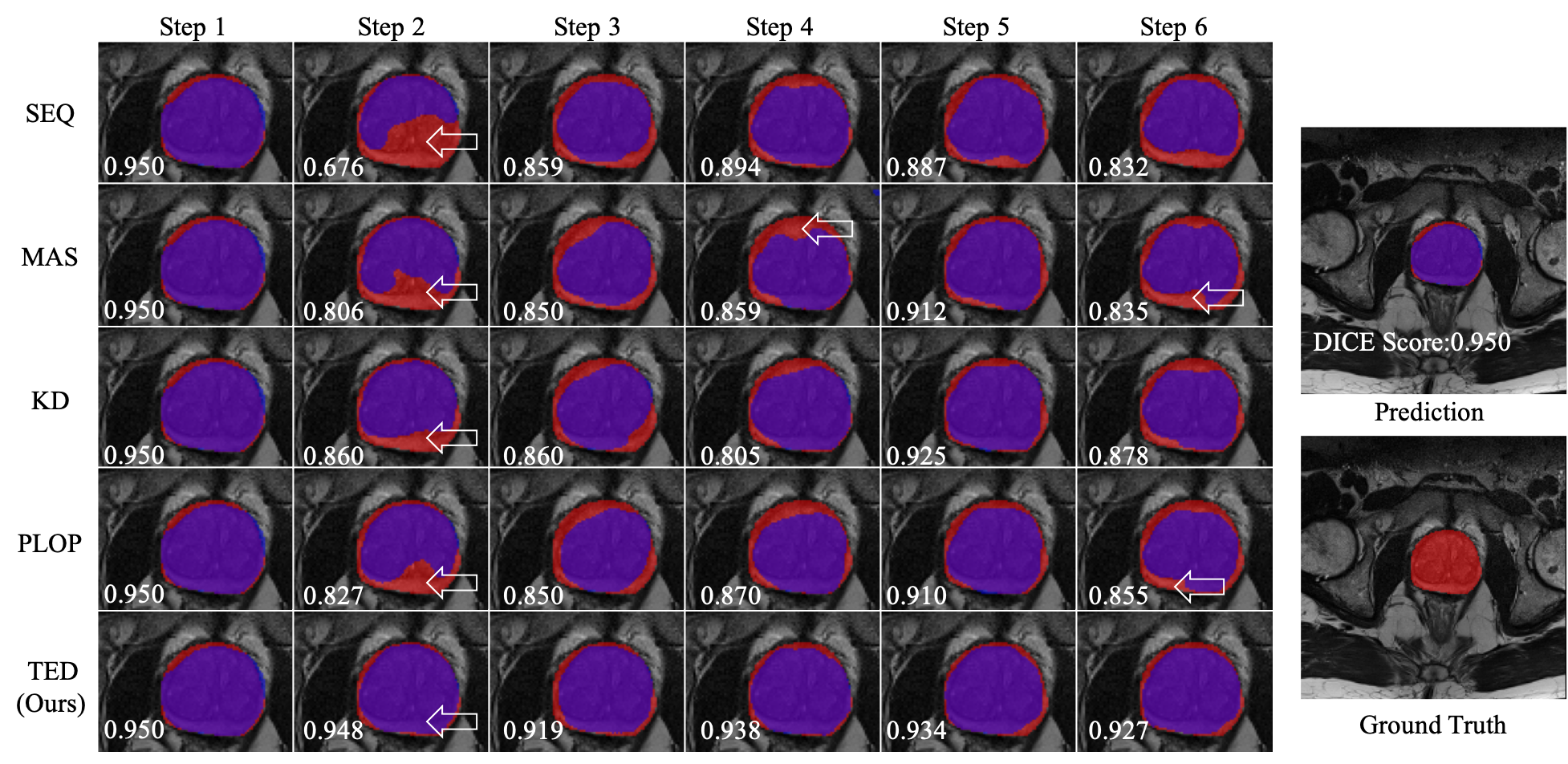}
	\caption{The process of catastrophic forgetting within the initial domain, RUNMC, highlights TED's resilience against forgetting, standing out distinctly when compared with other methods in the realm of domain continual medical image segmentation.} 
	\label{prostate_continual}
\end{figure*} 

\begin{figure*}[t]
	\centering
	\includegraphics[width=0.98\textwidth]{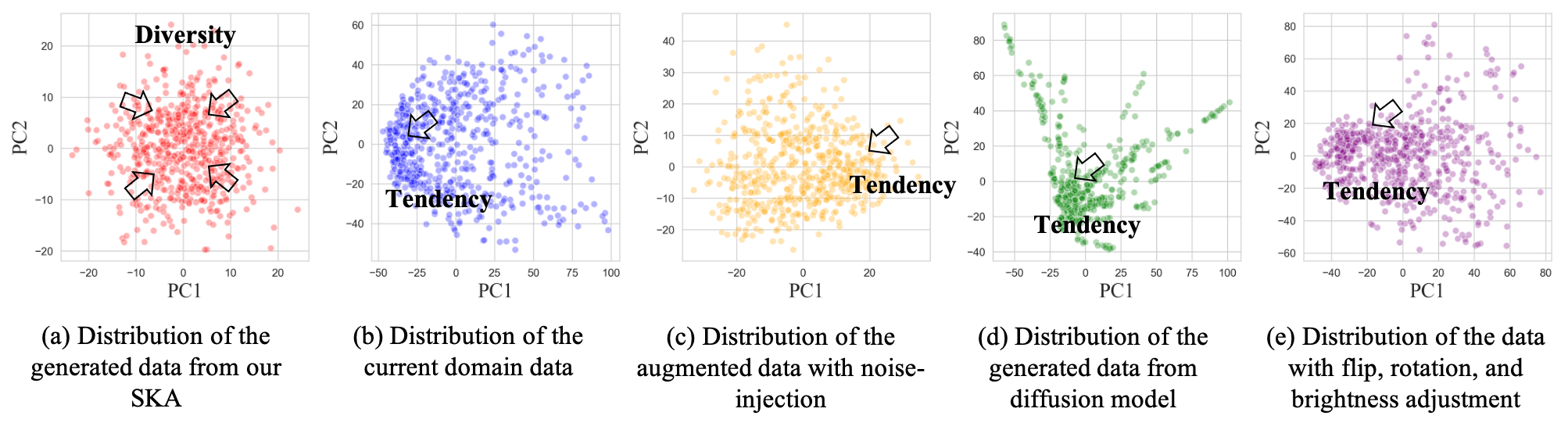}
	\caption{Our stochastic knowledge augmentation (SKA) showcases superior knowledge diversity (data distribution) compared to other methods on the prostate dataset. This diversity is evident in the features extracted from the bottleneck layer of the segmentation network, which are compressed to two values via PCA. Specifically:(a) Our SKA presents uniformly diverse knowledge. (b) Old knowledge, when distilled by current domain data, manifests a tendency and is constrained in its diversity. (c) The introduction of noise into images for the purpose of distilling old knowledge results in patterns that echo those observed in (b) with limited diversity. (d) Old knowledge distilled via a diffusion model displays restricted diversity and an pronounced tendency in its distribution. (e) The distribution of augmented data, processed through operations such as flipping, rotation, and brightness adjustment, exhibits tendencies.} 
	\label{ablation_diversity}
\end{figure*}

\textbf{Segmentation Visualization:} 
Fig. \ref{result_prostate} and \ref{result_cardiac} effectively illustrate the efficacy of our tri-enhanced distillation (TED) framework in addressing catastrophic forgetting, showcasing superior performance compared to other methods.
They exhibit the highest overlap ratio and the most precise results in comparison to other methods. Fig. \ref{prostate_continual} visualizes the process of forgetting for an instance from the Prostate dataset's initial domain, RUNMC.  It's evident that our approach consistently exhibits enhanced resistance to forgetting throughout the continual learning process. Across these illustrations, the ground truth is delineated in red, predictions by the model in blue, and overlapping regions manifest in purple. The methodologies are cataloged in the first column, with the DICE score being presented in white text.

$\bullet$(1) Fig. \ref{result_prostate} renders the prostate segmentation outcomes post-training on the most recent domain dataset, HK. Since the initial domain, RUNMC, shares a comparable image domain with HK, it retains commendable segmentation performance. Although the comparison methods showcase analogous segmentation results, our TED emerges superior with a DICE score of 91.5\%. Notably, the I2CVB dataset, vastly different from HK, presents challenges across methods. Nevertheless, TED maintains consistent performance. Subsequent domains, UCL and BIDMC, present intricate semantic segmentation issues, evident in methods like PLOP. However, TED persists in retaining semantic fidelity. Overall, TED manifests exemplary visualization-based segmentation results across all six domains.

$\bullet$(2) Fig. \ref{result_cardiac} offers a visualization into cardiac segmentation results on the left ventricular endocardium (LV-endo) and the right ventricle (RV), following training on the Canon dataset. For LV-endo segmentation, even though the initial dataset has been overshadowed by subsequent data, all methods adeptly segment it, attributable to its prominent structure. The GE domain, with its small region of interest, proves challenging; SEQ and KD tend to over-segment, whereas TED maintains optimal segmentation. Despite Canon's unique perspective, TED continues to deliver great results. For RV segmentation, TED's proficiency is evident, especially in the challenging GE domain. The visualizations accentuate TED's resilience against forgetting, compared with other methods.

$\bullet$(3) Fig. \ref{prostate_continual} displays the process of catastrophic forgetting in the initial domain, RUNMC. It shows TED's robust anti-forgetting prowess compared with other methods throughout domain continual medical image segmentation. The chosen example is straightforward, facilitating a clearer observation of post-training effects on each new domain. Upon training RUNMC at Step 1, all methods yield commendable results with a DICE score of 95.0\%. This uniformity in DICE is attributed to the consistent training configurations used by all methods on the RUNMC dataset. By Step 2, a discernible degradation is evident in all except TED, with SEQ and MAS particularly affected. As training progresses to Step 3, while there's a general improvement, TED stands out with a DICE of 91.9\%. By Step 4, except for KD, there's an improvement in performance, with TED registering a DICE of 93.8\%. Step 5 introduces minimal fluctuations. Finally, at Step 6, although there's a dip across methods, TED's results experience the least deterioration.

\begin{table*}[!h]
	\centering
	\small
	\caption{The ablation study on the cardiac dataset demonstrates  the contribution of our proposed components. Integrating the novel components into the naive knowledge distillation (KD) leads to a marked enhancement in performance. Conversely, removing them from our TED results in a decrease in performance.}
	\label{ablation_cardiac}
	\begin{tabular}{r|p{5mm}|p{5mm}|p{5mm}|p{5mm}||cc|cc|cc|cc|cc|cc}
		\toprule
		\#&\multicolumn{4}{c||}{components}&\multicolumn{4}{c|}{LV-endo}&\multicolumn{4}{c|}{LV-epi}&\multicolumn{4}{c}{RV}\\
		\midrule
		&\multicolumn{4}{c||}{}&\multicolumn{2}{c|}{DSC$\uparrow$}&\multicolumn{2}{c|}{HD95$\downarrow$}&\multicolumn{2}{c|}{DSC$\uparrow$}&\multicolumn{2}{c|}{HD95$\downarrow$}&\multicolumn{2}{c|}{DSC$\uparrow$}&\multicolumn{2}{c}{HD95$\downarrow$}\\
		&KD& SKA & AKT & GVU & AVG & BWT &AVG& BWT& AVG & BWT &AVG& BWT& AVG & BWT &AVG& BWT\\
		\midrule
		1&\checkmark&&&&0.853&-0.035&21.26&12.30&0.720&-0.091&19.14&11.54&0.805&-0.048&19.20&4.85\\
		2&\checkmark&\checkmark&&&{0.867}&{-0.014}&{15.00}&\textbf{2.71}&{0.733}&\textbf{-0.070}&22.56&{11.31}&{0.824}&{-0.017}&{16.18}&{-1.35}\\
		3&\checkmark&&\checkmark&& {0.878}&{-0.009}&{19.69}&{8.18}&{0.737}&{-0.078}&22.05&15.42&{0.802}&{-0.040}&\textbf{15.44}&{-0.01}\\
		4&\checkmark&&\checkmark&\checkmark& 0.879 & -0.006 & 18.31 & 6.11 & 0.724 & -0.095 & 19.04 & 13.42 & 0.814 &-0.037 & 19.41 & 5.23 \\
		5&\checkmark&\checkmark&\checkmark&&0.874&-0.014&16.58&3.24&0.720&-0.091&\textbf{18.20}&10.23&0.823&-0.030&17.36&1.16\\
		6&\checkmark&\checkmark&\checkmark&\checkmark&\textbf{0.887} &\textbf{0.015}&\textbf{13.91}&2.90&\textbf{0.735}&-0.075&20.60&\textbf{9.91}&\textbf{0.831}&\textbf{-0.012}&16.27&\textbf{-2.84}\\
		\bottomrule
	\end{tabular}
\end{table*}

\begin{table}[!h]
	\centering
	\small
	\caption{The ablation study on the prostate dataset demonstrates the contribution of our proposed components. Integrating the novel components into the naive knowledge distillation (KD) leads to a marked enhancement in performance. Conversely, removing them from our TED results in a decrease in performance.}
	\label{ablation_prostate}
	\begin{tabular}{r|p{5mm}|p{5mm}|p{5mm}|p{5mm}||p{6mm}|p{9mm}|p{6mm}|p{6mm}}
		\toprule
		\#&\multicolumn{4}{c|}{components}&\multicolumn{2}{c|}{DSC$\uparrow$}&\multicolumn{2}{c}{HD95$\downarrow$}\\
		&KD& SKA & AKT & GVU & AVG & BWT &AVG& BWT\\
		\midrule
		1&\checkmark&&&&0.709&-0.105&29.18&15.02\\
		2&\checkmark&\checkmark&&&  {0.754}&{-0.065} &{22.12}&{7.27}\\
		3&\checkmark&&\checkmark&&  {0.741}&{-0.078} &{30.37}&{16.36}\\
		4&\checkmark&&\checkmark&\checkmark&  0.744&-0.088 &29.94&17.40\\
		5&\checkmark&\checkmark&\checkmark&&0.753&\textbf{-0.056}&22.07&8.39\\
		6&\checkmark&\checkmark&\checkmark&\checkmark&\textbf{0.759} &-0.059&\textbf{21.44}&\textbf{6.72}\\
		\bottomrule
	\end{tabular}
\end{table}

\subsection{Ablation Study for Key Components}

In this subsection, we conduct an ablation study of the three key components within our tri-enhanced distillation (TED) framework: stochastic knowledge augmentation (SKA), adaptive knowledge transfer (AKT), and global uncertainty-guided fusion (GUGF). To start, we assess each component in isolation to underscore their individual efficacies. We then tabulate the results, emphasizing the performance variations observed when either incorporating or excluding each component. Subsequently, we explore the unique roles these components fulfill, illustrating their influences on diversity, accuracy, and stability using visual aids.

\textbf{(1) Stochastic knowledge augmentation for more diversity.}

\begin{table}[!h]
	\centering
	\small
	\caption{The study of noise level on stochastic knowledge augmentation (SKA). The results show that different noise levels have a minimal impact on the model's average performance (AVG). In our default option, SKA chooses 0.0 mean and 1.0 standard deviation.}
	\label{noise_level}
	\begin{tabular}{l|cc|cc|cc}
		\toprule
		\multirow{4}{8pt}{Standard Deviation}&\multicolumn{6}{c}{Mean}\\
		\cmidrule(l){2-7}
		{}&\multicolumn{2}{c|}{0.0}&\multicolumn{2}{c|}{-0.5}&\multicolumn{2}{c}{0.5}\\
		{}&\multicolumn{2}{c|}{standard}&\multicolumn{2}{c|}{negative}&\multicolumn{2}{c}{positive}\\
		\cmidrule(l){2-7}
		{}&AVG& BWT&AVG& BWT&AVG& BWT\\
		\midrule
		0.1 small&0.760&-0.048&0.711&-0.110&0.763&-0.059\\
		\midrule
		0.5 medium&0.759&-0.065&0.755&-0.064&0.758&-0.057\\
		\midrule
		1.0 large&0.754&-0.065&0.763&-0.063&0.759&-0.048\\
		\bottomrule
	\end{tabular}
\end{table}

Tables \ref{ablation_cardiac} and \ref{ablation_prostate} substantiate that our stochastic knowledge augmentation (SKA) significantly improves the average segmentation performance and enhances the model's resilience against catastrophic forgetting. As shown in the second row, we integrated SKA into the basic knowledge distillation method, resulting in a significant performance improvement on both the prostate and cardiac datasets. As shown in the fourth row, upon removing SKA from the tri-enhanced distillation framework, we observed a decrease in performance, highlighting SKA's critical role in improving mitigation of catastrophic forgetting.
\begin{figure*}[t]
	
	\centering
	\begin{subfigure}[b]{0.33\textwidth}
		\includegraphics[width=\textwidth]{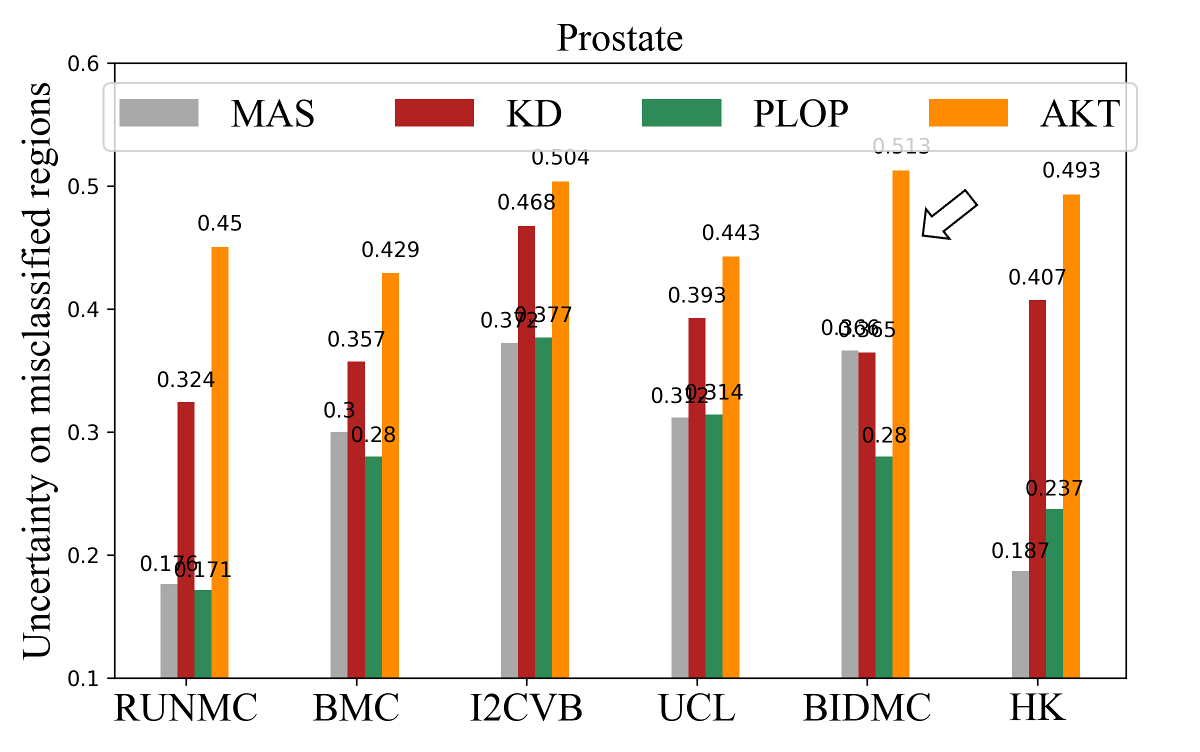}
		\caption{Prostate dataset}
		\label{ablation_accuracy_prostate}
	\end{subfigure}
	\hfill
	\begin{subfigure}[b]{0.66\textwidth}
		\includegraphics[width=\textwidth]{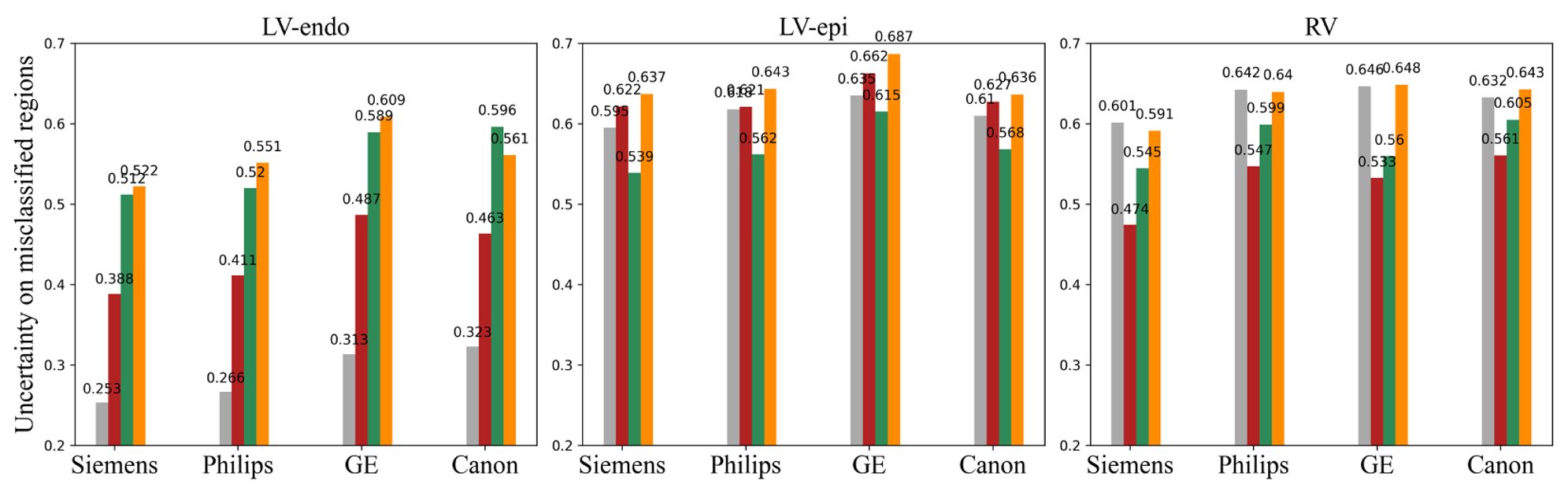}
		\caption{Cardiac dataset}
		\label{ablation_accuracy_cardiac}
	\end{subfigure}
	\caption{Our adaptive knowledge transfer (AKT) excels in capturing knowledge with heightened accuracy. The depicted bar plot provides a clear representation of uncertainty metrics within misclassified regions following training across all datasets. Incorporating our AKT yields an elevated average uncertainty in misclassified regions compared to other methods, signifying its superior proficiency in discerning crucial nuances associated with domain shifts.}
	\label{ablation_accracy}
\end{figure*}

Table \ref{noise_level} shows that different noise levels in our SKA have a minimal impact on the model's average performance (AVG). It displays three levels of noise standard deviation (0.1 small, 0.5 medium, 1.0 large) and three noise means (0.0 standard, -0.5 negative skew, 0.5 positive skew), measuring the model's average performance (AVG) and backward transfer (BWT) based on the DICE coefficient. We can observe notable variations in backward transfer (BWT) across different noise levels, especially under negative skew noise (-0.5 mean), where the BWT significantly decreases, particularly with small noise (0.1 standard deviation). This suggests that specific characteristics of noise, such as the direction of skewness, can impact the model's stability. Although performance can fluctuate, using the standard normal distribution is still a great default option.

Fig. \ref{ablation_diversity} visually illustrates the distribution of old knowledge, underscoring the superior diversity exhibited by SKA relative to other methods. This distribution is derived from features extracted from the bottleneck layer of a pre-trained segmentation network, which are then compressed into a two-dimensional space using Principal Component Analysis.
Fig. \ref{ablation_diversity} (a) demonstrates a uniform and diverse knowledge distribution, indicating a completed review of old knowledge by our SKA. 
In contrast, Fig. \ref{ablation_diversity} (b) and (c) showcase the knowledge distribution derived from current data and data augmented with noise-injection, respectively. Both exhibit similar patterns, indicating a redundancy of knowledge as highlighted in Issue 1 of Fig. \ref{contributions} (a). This repetition impedes the distillation process from achieving a completed review of old knowledge.
Fig. \ref{ablation_diversity} (d) showcases the distribution of old knowledge obtained through images generated by diffusion models. Interestingly, it exhibits an extremely tendentious distribution and underperforms compared to our SKA. This discrepancy may be due to the challenges diffusion models face when generating high-quality images from a limited number of data. Yet, our SKA effectively sidesteps this tendency-based distribution to mitigate the repetitive knowledge that emerges when the quality of generated images is substandard. Fig. \ref{ablation_diversity} (e) shows the distribution of old knowledge augmented by flip, rotation, and brightness adjustment. These operations are common in medical image augmentation. However, this is an enhancement in the domain where the new data image is located, and the knowledge distribution will still be tendentious. As a result, SKA not only amplifies the volume but also enhances the diversity of distilled knowledge.

\textbf{(2) Adaptive knowledge transfer for more accuracy.}

Tables \ref{ablation_cardiac} and \ref{ablation_prostate} also demonstrate the significant enhancements that our adaptive knowledge transfer (AKT) brings to the naive knowledge distillation (KD) method. Upon examining the prostate segmentation results presented at the third row in Table \ref{ablation_prostate}, it becomes evident that the incorporation of AKT yields a significant enhancement. Specifically, the average DICE score rises from 70.9\% to 74.1\%. In the cardiac segmentation results showcased at the third row in Table \ref{ablation_cardiac}, the overall DICE score across structures witnesses a significant improvement, underscoring the effectiveness of our AKT. While the HD95 metric for the left ventricular epicardium (LV-epi) indicates a marginal decline, it's worth noting that this might be due to the complex annular structure of the LV-epi, which inherently poses challenges for segmentation. Given the substantial benefits observed in the DICE scores, the minor setback in HD95 for LV-epi can be considered negligible. For the LV-endo structure, the average DICE improves from 85.3\% to 87.8\%. Similarly, for the RV struc ture, the average BWT sees a significant boost, moving from 4.85 to a nearly negligible -0.01, further underscoring the advantages of integrating AKT. 

To gauge the accuracy of knowledge capture, we employ uncertainty statistics within misclassified regions; higher uncertainty corresponds to more accurate knowledge. Our choice to utilize uncertainty as a metric for quantifying the accuracy of knowledge is grounded in the fact that uncertainty serves as a robust indicator of domain shifts within data distributions \citep{mehrtash2020confidence}. When a domain shift occurs, a model should predict heightened uncertainty. Hence, when predicting on datasets from old domains, the model should manifest increased uncertainty in misclassified regions, which further signifies the accuracy of its knowledge capture. Additionally, we measure uncertainty by calculating the entropy of the output probability, expressed as: $\mathcal{H} = - (plogp + (1-p)log(1-p)) / log2$, where $p$ represents the probability.

Fig \ref{ablation_accuracy_prostate} and \ref{ablation_accuracy_cardiac} accentuate the prowess of AKT in capturing old knowledge with enhanced accuracy. As illustrated, the yellow colorbar in our AKT method consistently towers above those of other methods. Post-training on the final domain dataset, we have collated the uncertainty statistics of misclassifications across both old and current datasets for various methods. For prostate segmentation (Fig \ref{ablation_accuracy_prostate}), our approach boasts an average uncertainty of 0.5 across diverse datasets, surpassing other methods like KD, which averages around 0.3. In cardiac segmentation (Fig \ref{ablation_accuracy_cardiac}), our method consistently achieves the highest uncertainty across three distinct anatomical structures. While some alternate methods also register high uncertainty, their performance varies significantly across different anatomical structures. For instance, while MAS achieves uncertainty close to our AKT for RV, its statistics for LV-endo considerably lag behind our method. The uncertainty statistics for both prostate and cardiac structures provide compelling evidence of the superior capability of our proposed AKT in capturing accurate knowledge.

\begin{table}[!h]
	\centering
	\small
	\caption{Ablation analysis of the global-view uncertainty (GVU) within our global uncertainty-guided fusion (GUGF) on the prostate dataset. The results indicate that the GVU's inclusion neither degrades nor significantly boosts segmentation performance. Performance deltas upon GVU's integration into AKT or its omission from TED remain subtle. The \textbf{$\Delta$} represents the absolute difference after adding the GVU, and the \textbf{$*$} highlights the improvement.}
	\label{ablation_prostate_bbu}
	\begin{tabular}{c||cc|cc}
		\toprule
		\multirow{2}{*}{}&\multicolumn{2}{c|}{DSC$\uparrow$}&\multicolumn{2}{c}{HD95$\downarrow$}\\
		&AVG& BWT&AVG& BWT\\
		\midrule
		AKT &{0.741}& {-0.078} &{30.37}&{16.355}\\
		AKT + GVU &  0.744&-0.088 &29.94&17.40\\
		$\Delta$ & {0.003$*$} & {0.010}& {0.43$*$} & {1.05}\\
		\midrule
		TED w/o GVU&0.753&{-0.056}&22.07&8.39\\
		TED (Ours)&0.759 &-0.059&21.44&6.72\\
		$\Delta$ & {0.006$*$}&  {0.003}& {0.63$*$}& {1.67$*$}\\
		\bottomrule
	\end{tabular}
\end{table}
\begin{table*}[!t]
	\centering
	\small
	\caption{Ablation analysis of the global-view uncertainty (GVU) within our global uncertainty-guided fusion (GUGF) on the cardiac dataset. The study reveals minimal degradation in segmentation performance upon its inclusion. The integration of GVU into AKT and its exclusion from TED result in nuanced performance shifts. The \textbf{$\Delta$} represents the absolute difference after adding the GVU, and the \textbf{$*$} highlights the improvement.}
	\label{ablation_cardiac_bbu}
	\begin{tabular}{|c||cc|cc|cc|cc|cc|cc|}
		\toprule
		{Structures}&\multicolumn{4}{c|}{LV-endo}&\multicolumn{4}{c|}{LV-epi}&\multicolumn{4}{c|}{RV}\\
		\midrule
		\multirow{2}{*}{}&\multicolumn{2}{c|}{DSC$\uparrow$}&\multicolumn{2}{c|}{HD95$\downarrow$}&\multicolumn{2}{c|}{DSC$\uparrow$}&\multicolumn{2}{c|}{HD95$\downarrow$}&\multicolumn{2}{c|}{DSC$\uparrow$}&\multicolumn{2}{c|}{HD95$\downarrow$}\\
		&AVG& BWT&AVG& BWT&AVG& BWT&AVG& BWT&AVG& BWT&AVG& BWT\\
		\midrule
		AKT& 0.878&-0.009&19.69&8.18&0.737&-0.078&22.05&15.42&0.802&-0.040&15.44&-0.01\\
		AKT + GVU & 0.879 & -0.006 & 18.31 & 6.11 & 0.724 & -0.095 & 19.04 & 13.42 & 0.814 &-0.037 & 19.41 & 5.23 \\
		$\Delta$ &{0.001$*$}&{0.003$*$}&{1.68$*$}&{1.07$*$}&{0.013}&{0.017}&{3.01$*$}&{2.00$*$}&{0.012$*$}&{0.003$*$}&{ 3.97}&{5.24}\\
		\midrule
		TED - GVU &0.874&-0.014&16.58&3.24&0.720&-0.091&18.20&10.23&0.823&-0.030&17.36&1.16\\
		TED (ours)&0.887 &0.015&13.91&2.90&0.735&-0.075&20.60&9.91&0.831&-0.012&16.27&-2.84\\
		$\Delta$ &{0.013$*$}&{0.029$*$}&{2.67$*$}&{0.34$*$}&{0.015$*$}&{0.016$*$}&{2.40}&{0.3$*$}&{0.002$*$}&{0.018$*$}&{1.09$*$}&{4.00$*$}\\
		\bottomrule
	\end{tabular}
\end{table*}

\textbf{(3) Global uncertainty-guided fusion for more stability.}
\begin{figure*}[!h]
	
	\centering
	\begin{subfigure}[b]{0.96\textwidth}  
		\includegraphics[width=\textwidth]{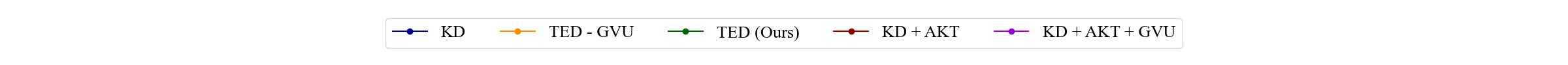}
	\end{subfigure}
	\begin{subfigure}[b]{0.24\textwidth}
		\includegraphics[width=\textwidth]{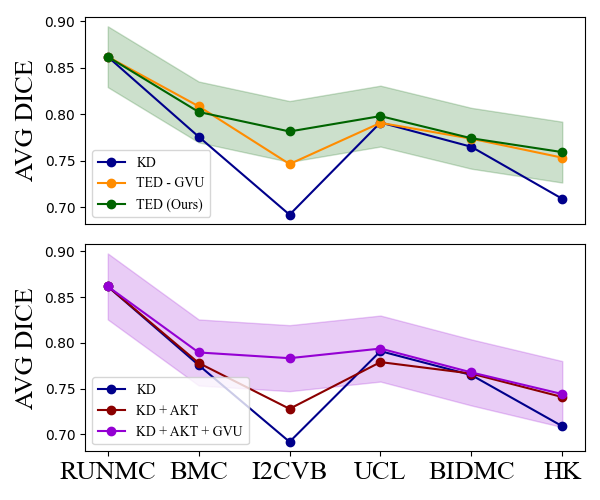}
		\caption{AVG DICE on prostate segmentation}
		\label{fig:sub1}
	\end{subfigure}
	\hfill
	\begin{subfigure}[b]{0.24\textwidth}
		\includegraphics[width=\textwidth]{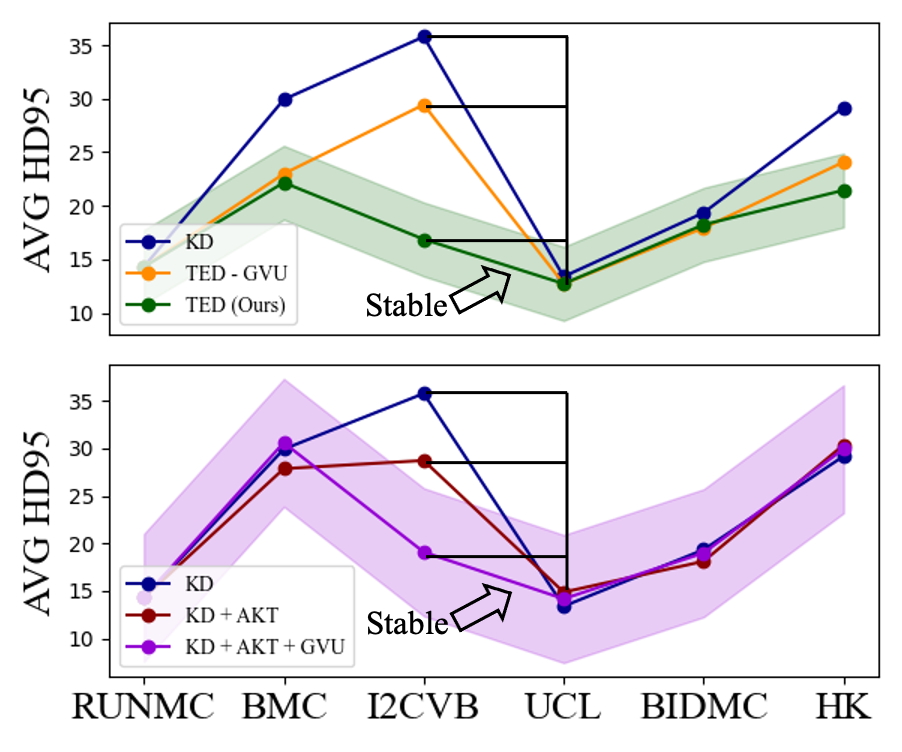}
		\caption{AVG HD95 on prostate segmentation}
		\label{fig:sub2}
	\end{subfigure}
	\hfill
	\begin{subfigure}[b]{0.24\textwidth}
		\includegraphics[width=\textwidth]{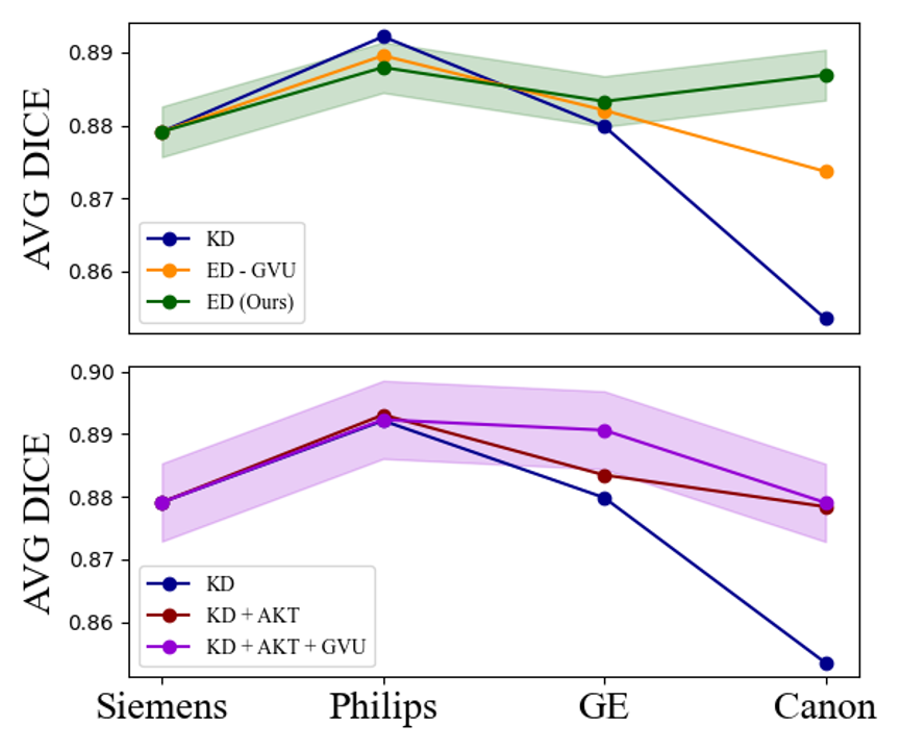}
		\caption{AVG DICE on LV-endo segmentation}
		\label{fig:sub3}
	\end{subfigure}
	\hfill
	\begin{subfigure}[b]{0.24\textwidth}
		\includegraphics[width=\textwidth]{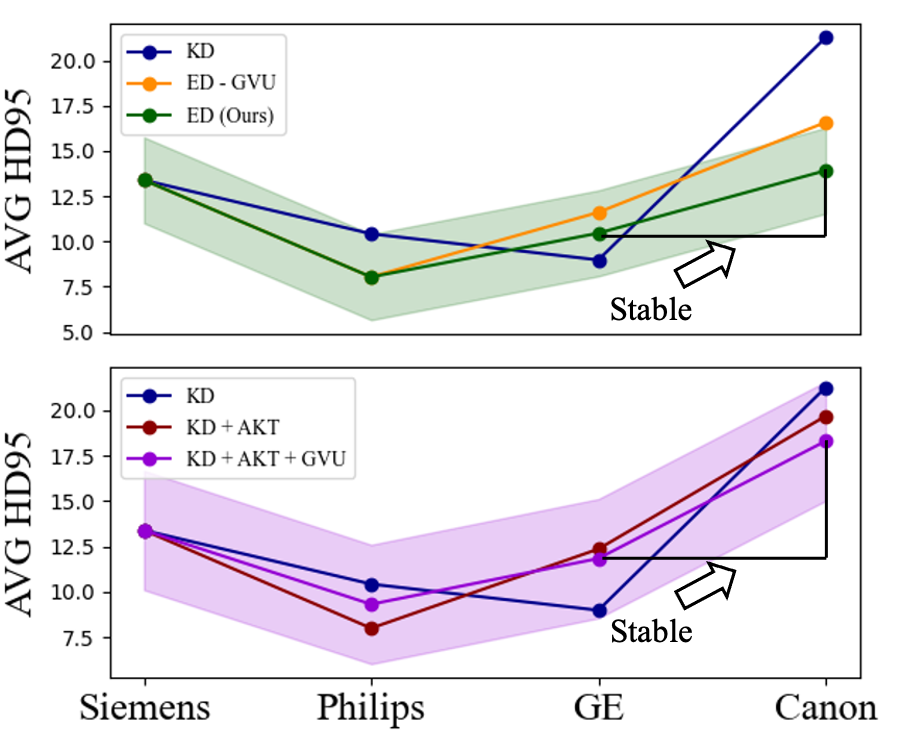}
		\caption{AVG HD95 on LV-endo segmentation}
		\label{fig:sub4}
	\end{subfigure}
	\caption{Evolution of average DICE and HD95 scores with the introduction of new data. The figure highlights the superior stability of our global uncertainty-guided fusion (GUGF) with a global-view uncertainty (GVU). Stability is gauged by the consistency in the model's performance, with minimal fluctuations indicating optimal stability during domain continual medical image segmentation. The performance curves of 'TED - GVU' and 'KD + AKT + GVU' are juxtaposed against the baseline 'KD', serving as comparative references to elucidate the stabilizing influence of GVU.}
	\label{ablation_stability}
\end{figure*}

Fig. \ref{ablation_stability} depicts the process of average DICE and HD95 as new data becomes available. Our global uncertainty-guided fusion (GUGF), incorporating the perspective of global-view uncertainty (GVU), showcases optimal stability throughout. Good stability is characterized by minimal fluctuation in the model's performance during domain continual medical image segmentation.
On one hand, we observe the stability variations in performance curves when GVU is removed from our TED (TED - GVU). On the other hand, we use naive knowledge distillation as a baseline to observe the changes in performance curve stability after introducing GVU on AKT (KD + AKT + GVU).
In Fig. \ref{fig:sub1} and \ref{fig:sub2}, which present the DICE and HD95 changes for prostate segmentation, the stability is evidently superior when GVU is integrated. For instance, after training on the I2CVB dataset, the method without GVU exhibits significant performance jitters. In contrast, the method with GUV (TED and KD + AKT+ GVU) remains considerably stable.
Similarly, Fig. \ref{fig:sub3} and \ref{fig:sub4}, which display the DICE and HD95 variations for cardiac LV-endo segmentation, reinforce the improved stability achieved by integrating GVU. For example, upon completing training with the final domain, Canon, the approach incorporating GUV consistently maintains stability.

As evidenced by Tables \ref{ablation_prostate_bbu} and \ref{ablation_cardiac_bbu}, the introduction of the Global-View Uncertainty (GVU) exerts minimal impact on the overall performance. This ablation study was specifically designed to ascertain that the integration of GVU doesn't compromise the performance of the TED framework. To circumvent potential interference from the SKA module in our TED, we conducted ablation experiments with both the presence (TED-GVU) and absence (KD+AKT+GVU) of SKA. From Table \ref{ablation_prostate_bbu}, it's evident that the incorporation of GVU affects the average DICE score by less than 1.0\%. Similarly, its influence on the HD95 metric remains below 1, and its bearing on the BWT is also inconsequential. Parallel observations can be drawn from Table \ref{ablation_cardiac_bbu} for the three anatomical structures of the heart. Across these structures, the impact of GVU remains consistently minimal, underscoring its non-intrusiveness within the segmentation process.

\section{Conclusion}
Our proposed tri-enhanced distillation (TED) framework provides a powerful solution for improving the mitigation of catastrophic forgetting in domain continual medical image segmentation. To address the issues at three stages of existing distillation-based methods, this innovative framework effectively boosts knowledge diversity, accuracy, and stability. This framework incorporates three key components: stochastic knowledge augmentation (SKA), adaptive knowledge transfer (AKT), and global uncertainty-guided fusion (GUGF). Together, SKA, AKT, and GUGF form a potent combination, driving the effectiveness of our TED framework. Our experimental results confirm the effectiveness of our approach and demonstrate superior performance compared to state-of-the-art methods. We are convinced that our TED framework will set a strong benchmark in the field of domain continual medical image segmentation. In the future, we will explore domain specific augmentation approaches for domain continual medical image segmentation.

\section*{Acknowledgments}
This work was supported by the National Natural Science
Foundation of China under Grant 62372135, Grant 62272135 and Grand 62202092

\bibliographystyle{model2-names.bst}\biboptions{authoryear}
\bibliography{reference}



\end{document}